\pdfoutput=1
\documentclass[conference]{sig-alternate-10pt}

\usepackage{booktabs} 
\usepackage[ruled]{algorithm2e} 

\usepackage{color}
\usepackage{verbatim}
\usepackage{array, calc}
\usepackage{xspace}
\usepackage{enumerate}
\usepackage{graphics, epstopdf, graphicx, epsfig, psfrag, epsf}
\usepackage{amssymb, amsfonts, amsmath, times}
\usepackage{multicol, multirow, balance, verbatim, caption}
\usepackage{subcaption}
\usepackage[mediumspace,mediumqspace,Grey,squaren]{SIunits}
\usepackage{tabularx}
\usepackage{authblk, ragged2e}

\usepackage{amsmath, amsthm, amssymb}

\usepackage{relsize}

\usepackage{url}
\usepackage{breakurl}

\newcommand{\ie}{{\em i.e., }}
\newcommand{\eg}{{\em e.g., }}

\usepackage{xcolor}
\definecolor{heraldBlue}{rgb}{0.0,0.0,0.8}
\definecolor{heraldGreen}{rgb}{0.0,0.4,0.0}
\definecolor{heraldPurple}{rgb}{0.9,0.1,0.9}

\newcommand{\amnote}[1]{\textcolor{heraldRed}{ #1}}

\newcommand{\blue}[1]{\textcolor{blue}{#1}}
\newcommand{\red}[1]{\textcolor{red}{#1}}

\newcommand\textvtt[1]{{\sffamily\relsize{-.6}\selectfont #1}}

\newcommand{\antshield}{\textvtt{Ant\-Shield}\xspace}
\newcommand{\recon}{\textvtt{Recon}\xspace}

\newcommand{\querykeys}{\textvtt{query\-~keys}\xspace}
\newcommand{\fscore}{F1\-~score\xspace}
\newcommand{\standalone}{\textvtt{Local}\xspace}
\newcommand{\federated}{\textvtt{Fe\-de\-ra\-ted}\xspace}
\newcommand{\centralized}{\textvtt{Cen\-tra\-lized}\xspace}
\newcommand{\even}{\textvtt{e\-ven}\xspace}
\newcommand{\uneven}{\textvtt{un\-even}\xspace}
\newcommand{\Even}{\textvtt{E\-ven}\xspace}
\newcommand{\Uneven}{\textvtt{Un\-even}\xspace}
\newcommand{\nomoads}{\textvtt{NoMo\-Ads}\xspace}

\newcommand{\filerequest}{\textvtt{file\-~request}\xspace}
\newcommand{\svm}{SVM\xspace}
\newcommand{\svms}{SVMs\xspace}
\newcommand{\uri}{URI\xspace}
\newcommand{\cookie}{Coo\-kie\xspace}
\newcommand{\webview}{\textvtt{Web\-view}\xspace}
\newcommand{\reconstyle}{\textvtt{Re\-con\-~Words}\xspace}

\newcommand{\keyless}{\textvtt{key\-less}\xspace}
\newcommand{\Keyless}{\textvtt{Key\-less}\xspace}
\newcommand{\allwords}{\textvtt{All\-~Words}\xspace}
\newcommand{\httpkeys}{\textvtt{HTTP\-~Keys}\xspace}
\newcommand{\inhouse}{\textvtt{in-\-house}\xspace}
\newcommand{\Inhouse}{\textvtt{In-\-house}\xspace}

\newcommand{\FOR}{\textbf{for }}

\newcommand{\DO}{\textbf{ do }}

\newcommand{\RETURN}{\textbf{return }}

\newcommand{\SP}{${}$\hspace{0.25cm}}

\newfont{\mycrnotice}{ptmr8t at 7pt}
\newfont{\myconfname}{ptmri8t at 7pt}
\let\crnotice\mycrnotice%
\let\confname\myconfname%

\begin{document}
	
\title{A Federated Learning Approach for\\  Mobile Packet Classification}

	\author{Evita Bakopoulou}


	\author{B\'alint Tillman}

	
	\author{Athina Markopoulou}
	\affil{University of California, Irvine}
	\affil{{\{{ebakopou, tillmanb, athina}\}@uci.edu}}

\maketitle

\begin{abstract}  	
	{In order to improve mobile data transparency, a number of network-based approaches have been proposed to inspect packets generated by mobile devices and detect personally identifiable information (PII), ad requests, or other activities. State-of-the-art approaches train classifiers based on features extracted from HTTP packets. So far, these classifiers have only been trained in a centralized way, where mobile users label and upload their packet logs to a central server, which then trains a global classifier and shares it with the users to apply on their devices. However, packet logs used as training data may contain sensitive information that users may not want to share/upload. In this paper, we apply, for the first time, a Federated Learning approach to mobile packet classification, which allows mobile devices to collaborate and train a global model, without sharing raw training data.  Methodological challenges we address in this context include: model and feature selection, and tuning the Federated Learning parameters.  We apply our framework to two different packet classification tasks (\ie to predict PII exposure or ad requests in HTTP packets) and we demonstrate its effectiveness in terms of classification performance, communication and computation cost, using three real-world datasets.}
\end{abstract}

\section{Introduction}\label{sec:introduction}

\begin{figure*}[t]
	\centering
	\begin{subfigure}{0.235\textwidth}
		\centering
		\includegraphics[width=0.99\linewidth]{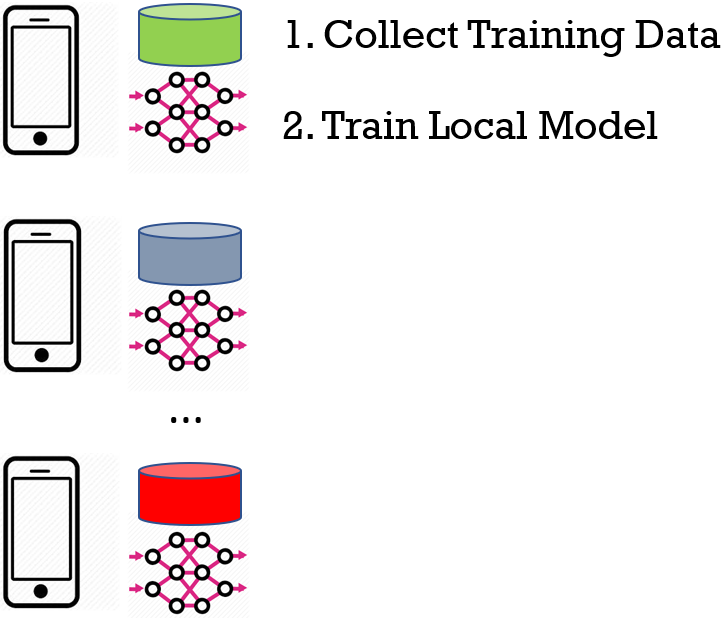}
		\caption{Local: share nothing}
	\end{subfigure}	
	\begin{subfigure}{0.305\textwidth}
		\centering
		\includegraphics[width=0.99\linewidth]{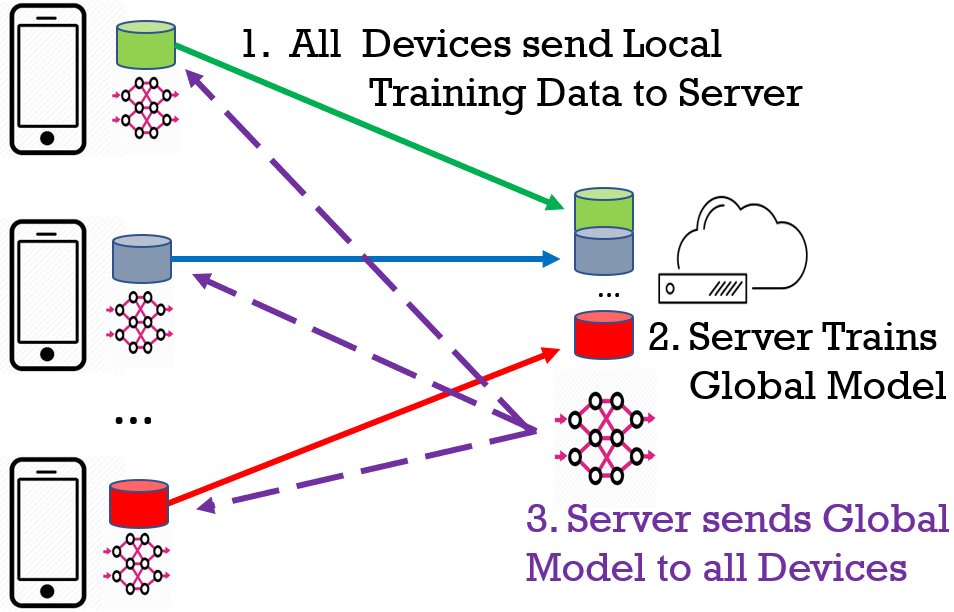}
		\caption{Centralized: devices share training data}
	\end{subfigure}	
	\begin{subfigure}{0.37\textwidth}
		\centering
		\includegraphics[width=0.99\linewidth]{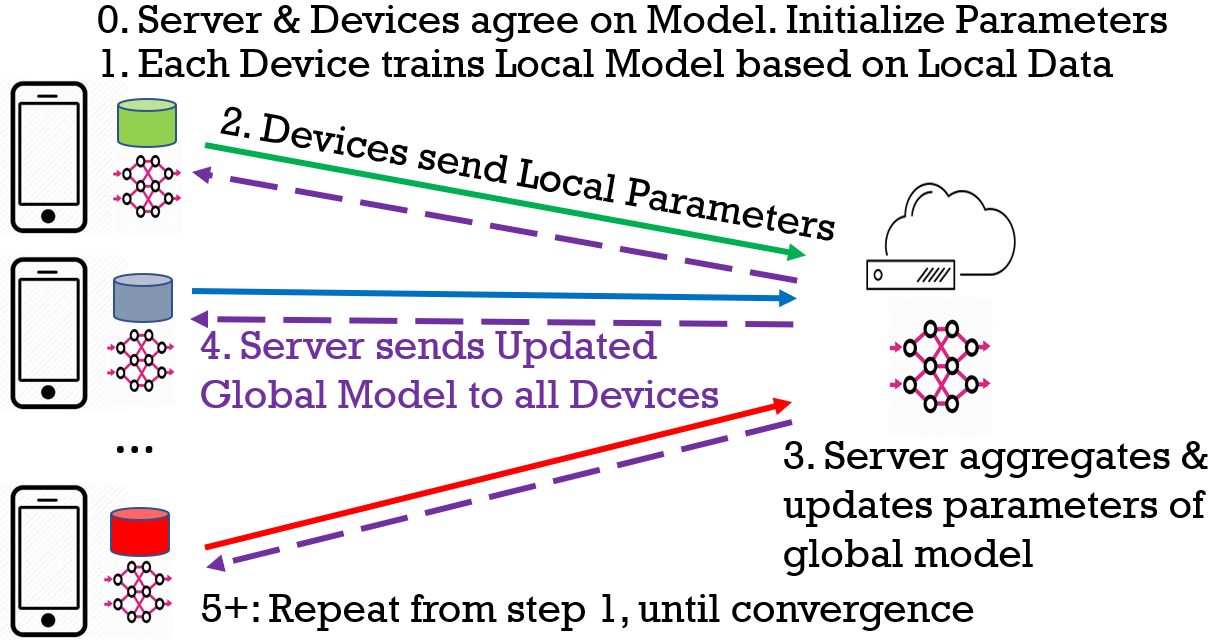}
		\caption{Federated: Devices share model parameters, not data} 
	\end{subfigure}	
	
	\caption{Overview of general approaches to train machine learning models for packets from mobile devices.}
	
	\label{fig:distributed_learning_figure_v2}
\end{figure*}

\begin{figure}[t!]
	\centering
	\includegraphics[width=0.99\linewidth]{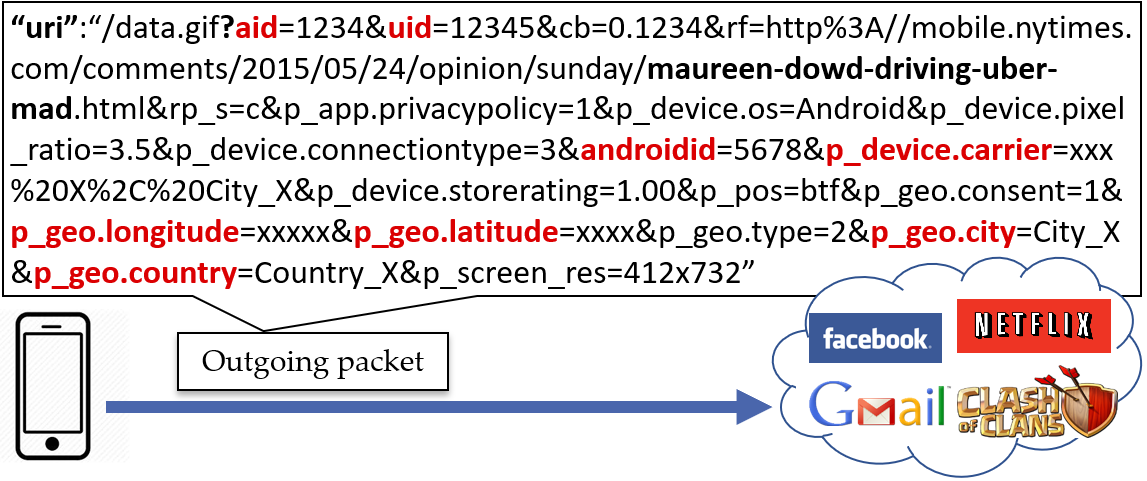}
	\caption{Example of an outgoing HTTP packet, sent from an app on the mobile device  to a remote server. The \uri field alone reveals a lot of information, including various identifiers, referred domain, location and other parameters that can be used for fingerprinting and tracking users.}
	\label{fig:user_profile}
\end{figure}




There is recently increased public awareness and concern about how sensitive information available on mobile devices  is shared and tracked. In particular, mobile apps and third party libraries (including developer, tracking and advertising libraries) routinely 
send such information (\ie personally identifiable information or ``PII'', sensory data, user activity) to remote servers, typically without the user being aware or in control of this information flow. Some steps are being taken to increase mobile, and more generally online, data transparency on the legal side (EU GDPR \cite{GDPR}, CCPA \cite{ccpa}, COPPA \cite{coppa}) as well as on the technical side, including approaches like permissions \cite{almuhimedi2015your, panoptispy, kratos, backes2016demystifying}, static and dynamic approaches \cite{gordon2015information, gordon2015droidsafe, arzt2014flowdroid, gibler2012androidleaks, pedal, jia2017open, taintdroid, hornyack2011appfence}, and network-based approaches \cite{recon15, antshield-arxiv, nomoads}.

In this paper, we focus on network-based approaches that  inspect packets transmitted out of mobile devices in order to detect PII, tracking, ad requests, malware or other activities; an example is depicted on Fig. \ref{fig:user_profile}.  This information can then be used to take action (\eg block outgoing packets), 
to inform the user or for measurement studies. Early approaches (such as Haystack/Lumen \cite{haystack-webpage, haystack_paper} and AntMonitor \cite{antmonitor15, antmonitor-arxiv, antmonitor-app}) performed deep packet inspection (DPI)  and string matching to find PII. Mobile browsers \cite{adblock_browser} use manually-curated filter-lists (such as EasyList \cite{easylists}, hpHosts \cite{hpHosts}, and AdAway \cite{adawayHosts}) to match URI and other information and block ad requests. 

Machine learning approaches have been recently proposed to predict PII  (\eg \recon \cite{recon15} and \antshield \cite{spawc, antshield-arxiv}) or ad requests (\nomoads \cite{nomoads}) in outgoing packets, based on features extracted from HTTP requests.
 These classifiers are trained using labeled packet traces, obtained either through manual/automatic testing of apps, or by devices 
labeling packets from real user activity to contribute them to a server. 
However, training has only been done in a \centralized way so far. 

Consider the scenario and options presented on Fig. \ref{fig:distributed_learning_figure_v2}.  In the \standalone approach (Fig. \ref{fig:distributed_learning_figure_v2}(a)), mobile users could label packets on their device, train and apply their own classifier locally. In this case, users do not share any information with untrusted servers or other users, thus preserving their privacy. However, they do not benefit from the global training data that is available on a large number of devices either.  In the \centralized approach (Fig. \ref{fig:distributed_learning_figure_v2}(b)), mobile users label and upload their packet logs to a central server, which then trains a global classifier and shares it with all users to apply on their devices. However, packet traces labeled with PII, tracking and advertising, contain sensitive information that users may not want to share with a server or other users. 
 In this paper, we propose Federated  Mobile Packet Classification, which combines the best of both worlds: it allows mobile devices to collectively train a global model, without sharing their raw training data that may contain sensitive information  (in the label, features, or any part of the HTTP packet).

The Federated Learning framework was proposed in \cite{original_federated, mcmahan2017federated}, to collaboratively train models for mobile devices without sharing raw training data.  An overview is depicted on Fig. \ref{fig:distributed_learning_figure_v2}(c). The main idea is that mobile devices train a local model, and send only model parameters to the server, instead of the raw training data; the server aggregates the information from all users, and sends the updated parameters of the global model to all devices, and the process repeats until convergence.  In this paper, we apply Federated Learning to classify outgoing HTTP packets w.r.t. two specific tasks, namely predicting whether an outgoing packet contains (1) a PII (which we refer to as PII exposure) or  (2)  an Ad request (which typically results in an ad being served in the HTTP response).  
Methodological challenges we had to address in order to apply Federated Learning to packet classification include: model and feature selection, and tuning the Federated Learning parameters. We also evaluated our methodology using three real-world datasets and showed that it achieves high classification \fscore for both classification tasks (PII exposure and Ad Request), with minimal computation and communication cost. 
This paper makes the following three contributions. 

 {\em 1. Feature Space for HTTP packets.} We propose a 
 feature space based on HTTP features that performs well for both classification tasks (since PII exposure and Ad requests use the same fields to profile users), while protecting sensitive information and reducing the feature space. 
First, we observe that not only training data, but also features can  expose sensitive information;  \eg that would be the case if some of the PII shown on Fig. \ref{fig:user_profile} were selected as features. 
 Therefore, we use 
 only \httpkeys as features from an HTTP packet, 
  (i) keys from \uri and \cookie fields (ii) custom HTTP headers and (iii) the presence of a file request. 
  We purposely do {\em not} use neither destination domains or hostnames, nor any information from the URI path (which could be sensitive itself if a user visits a sensitive website with \ie political, medical or religious content) but only the keys mentioned above. Prior work such as \recon \cite{recon15} and \nomoads \cite{nomoads} used all the words from the HTTP packets after discarding the most frequent ones and the rarest ones.
 Our choice of features not only minimizes the sharing of sensitive information, 
  but also reduces the number of parameters that need to be updated. 
  Second, we observe that the size of the feature space depends on the mobile apps and third-party libraries. For example, \webview apps can access any domain, which leads to an explosion of feature space size and wide variation across users; in contrast, non-\webview apps have a limited API and result in a small feature space, which is the same across different users.

{\em 2. Model Selection: Federated \svm.}  State-of-the-art classifiers for mobile packet classification have typically  trained Decision Trees (DT) 
to predict PII exposure (\recon \cite{recon15}) or Ad requests (\nomoads \cite{nomoads}) based on features extracted from HTTP packets. DT were chosen by prior work primarily for two reasons: (i) their good classification performance and (ii) their interpretability (nodes in the trees are intuitive rules). 
Unfortunately, DT do not naturally lend themselves to federation, which has been developed for models based on Stochastic Gradient Descent (SGD), primarily for Deep Neural Networks (DNNs) \cite{original_federated, mcmahan2017federated}. 
In this paper, we propose {\em Federated \svm} as the core of the federated packet classification framework. We show that  (i) 
Support Vector Machine (\svm)  performs similarly to DT for our problem, (ii) {\em Federated \svm} achieves similar \fscore to \centralized \svm, within few communication rounds and with low computation cost per user, and (iii) \svm can be as interpretable 
as DT and we also discuss knowledge transfer between the two.




 {\em 3. Evaluation using three real-world datasets and two classification tasks.} To evaluate our framework, we used  different datasets: the publicly available \nomoads for Ad requests \cite{nomoads, nomoads_data} and \antshield for PII exposures \cite{spawc, antshield_data}; and our \inhouse datasets with real users.  For the first two datasets, we create synthetic users by splitting the data evenly or unevenly, and we evaluate how it affects Federated Learning. We compare \federated to \centralized and \standalone approaches w.r.t. to the classification performance, communication rounds and computation.  We show that the \federated models are superior to \standalone models and comparable to their corresponding \centralized models, without requiring too many communication rounds or too much computation per client in order to achieve an \fscore above  0.90 for PII and above 0.84 for Ad request prediction. We also demonstrate the benefit of crowdsourcing: a relatively small number of users is sufficient to train a good \federated model that generalizes well to most users.

The rest of this paper is organized as follows. Section \ref{sec:related} reviews related work. Section \ref{sec:methodology} presents our methodology for federated packet classification. 
Section \ref{sec:dataset_description} describes the datasets used for evaluation. Section \ref{sec:results} presents results for various scenarios and provides insights along the way.  Section \ref{sec:conclusion} concludes the paper and discusses directions for future work.

\section{Related Work}\label{sec:related}

Performance measurements from mobile devices have captured the interest of many researchers over the years \cite{mobiperf, Gordon:2015:AMA:2742647.2742649, Mitigating2018, corner2017advertising, NICScatter, mobilyzer, MolaviKakhki:2015:ITD:2815675.2815691}.  There is also increasing interest in understanding and controlling PII exposure and user tracking  on mobile devices. Some proposed approaches include: permissions \cite{almuhimedi2015your, panoptispy, kratos, backes2016demystifying}, static analysis \cite{gordon2015information, gordon2015droidsafe, arzt2014flowdroid, gibler2012androidleaks, pedal, jia2017open}, dynamic analysis \cite{taintdroid, hornyack2011appfence}, and network-based approaches \cite{razaghpanah2018apps, ren2018bug, vallinarodriguez2016tracking}; our work falls within the latter. 


{\textbf{Interception of Mobile Traffic.} Network-based approaches inspect mobile traffic for PII exposure, or other information of interest, such as tracking, malware, advertising. 
		State-of-the-art tools include Haystack/Lumen \cite{haystack-webpage, haystack_paper, Razaghpanah2015HaystackAM}, Antmonitor \cite{antmonitor-demo, antmonitor-app, antmonitor-arxiv}. Lumen and Antmonitor use string matching to detect PII in outgoing packets sent from various apps to remote destinations. The interception of mobile traffic is not part of our contribution but it is orthogonal to our approach.}

\textbf{PII and Ad Detection via Blacklists.} There are many approaches based on manually curated blacklists \cite{easylists, hpHosts, adawayHosts} of domains on which they decide to block the whole packet destined to such domains or cookies from such domains \cite{garimella2017ad}. Since blacklists are hard to maintain due to the ever-changing advertising ecosystem, there are many efforts to update such blacklists with additional graph analysis \cite{vallina2016tracking}, or with machine learning \cite{bhagavatula2014leveraging, bau2013promising, gugelmann2015automated, nomoads}.

 {\bf Machine Learning-Based PII{\tiny } and Ad Detection.}  \recon \cite{recon15} and 
 \antshield \cite{antshield-arxiv} are machine learning approaches to detect PII exposure in outgoing HTTP packets: they train (offline, and in a centralized fashion) per-app/domain Decision Trees to detect PII exposures, based on features 
  extracted from HTTP packets. 
  \nomoads \cite{nomoads} is state-of-the-art approach for detecting Ad requests by enhancing blacklists via machine learning. We build on top of these ML approaches to introduce mobile packet classification learning in a distributed way. A step towards a more privacy-preserving PII detection on mobile devices is PrivacyProxy \cite{privacyproxy}, which processes user data locally and sends only hashed data to a server. One of its limitations is the cold start problem: it has to wait for enough data to be collected from other users in order to detect PII. MobiPurpose  \cite{mobipurpose} uses the keys only from network requests from apps, in addition to app meta-data and domain information, to classify the reason of PII exposure based on predefined candidates (\ie advertising, geo-tagging, etc.). All these approaches are \centralized, as they do not consider collaboration between users to leverage diverse app usage behaviors that can generate PII or Ad requests. We would like to note that our federated mobile classification approach can be used towards predicting other tasks, \ie fingerprinting detection \cite{bhat2018var, oh2017p, kurtz2016fingerprinting, zimmeck2017privacy, Korolova:2018:CTV:3176258.3176313, razaghpanah2018apps} assuming the availability of a corresponding per-packet labeled dataset. In this work, we focus on two classification tasks: PII exposure and Ad request detection because of the availability of labeled datasets that support these per-packet classification tasks.
 
  

{\bf Distributed Learning.} The authors in \cite{spawc} showed that systematic crowdsourcing where users collaborate with each other via data sharing helped to train better 
classifiers to detect PII exposures. However, it is assumed that users are willing to share their raw local data with a server and other users, which poses privacy risks.  In order to leverage crowdsourcing in a more privacy-preserving way, we considered two approaches to enable collaborative training of a global model from several mobile devices: Federated Learning (which is our focus in this paper) and Private Aggregation with Teacher Ensembles (PATE) \cite{Papernot2016}.  PATE trains a public Student model on labeled public data via a Teacher model trained on private data. In our problem, devices could train the Teacher, and public data could be 
datasets that have been made available by the research community or online communities, including those used in this paper \cite{nomoads_data, antshield_data}.  However, such public datasets do not necessarily capture all diverse patterns from apps, since they are produced via manual testing or automatic scripts that do not represent real users' app usage. Hence, we chose Federated Learning for our framework and problem space.  

{\bf Federated Learning.} An early version was proposed in \cite{Shokri:2015:PDL:2810103.2813687}, where users trained models locally and  shared the Stochastic Gradient Descent updates of certain parameters with a server, which then updated the global model. 
However \cite{Shokri:2015:PDL:2810103.2813687}  had no averaging mechanism and the evaluation was limited. The papers that coined the term ``Federated Learning'' were introduced 
in \cite{original_federated, mcmahan2017federated}, in order to train text and image classifiers using training data available  on a large number of mobile devices. The idea is that devices train SGD-based classifiers based on their local data and send updates (model parameters) to a trusted server, which aggregates them to update a global model.
The main advantage of this approach is that the raw training data does not leave  the device and thus, it is more privacy-preserving than a centralized model. 
A secondary advantage is that exchanging model parameters requires less communication (assuming fast convergence) than exchanging the raw training data, but this communication 
saving comes at some computational cost imposed on the devices to train models locally. Subsequent papers introduced optimizations in terms of communication efficiency, scalability and convergence \cite{konevcny2015federated, Konecny2016, Konecny2016_2, DBLP:journals/corr/abs-1812-07210, smith2017federated, guha2019one,  bonawitz2019towards, hard2018federated}, or optimization in client selection \cite{nishio2018client}. 

In contrast to related work in the field that is using image classification or next word prediction via word and character embeddings \cite{original_federated}, we focus on a problem where pre-trained word embeddings (such as Word2Vec \cite{word2vec}) are not applicable  due to non-dictionary words present in HTTP packets. We apply Federated Learning in a setting where shallow models' (such as SVMs) performance is comparable to state-of-the-art methods, this means that specialized deep learning model architectures (such as Convolutional Neural Networks (CNNs) or Long Short-Term Memory (LSTMs) \cite{original_federated}) are unnecessary.

\textbf{Privacy and Federated Learning.} 
Several security and privacy attacks are known for machine learning systems; \eg \cite{8049647, hitaj2017deep, Melis2018InferenceAA, melis2019exploiting,  metis, riazi2018deep} which include membership inference attacks, model inversion/extraction in ML models, poisoning the Federated model via malicious clients \cite{bhagoji2018analyzing, fung2018mitigating}, malicious server \cite{wang2018beyond} or training a backdoor task in addition to the main task in order to perform model replacement \cite{bagdasaryan2018backdoor}. Although Federated Learning protects the raw training data of each device and shares only model parameter updates, these updates (or even the features of the model \cite{melis2018}) may themselves leak information, due to privacy vulnerabilities of stochastic gradient descent (SGD) \cite{nasr2018comprehensive}. To prevent privacy attacks, additional mechanisms  have been proposed on top of Federated Learning, most notably, Secure Aggregation based on secure Multi-Party Computation (MPC)  \cite{fed_sec_aggregation} or Differential Privacy \cite{dwork2011differential} to offer privacy guarantees \cite{mcmahac_diff_priv, geyer2017, bhowmick2018protection, zhu2019federated} or both \cite{truex2018hybrid}. In this paper, we consider such variants of Federated Learning, as orthogonal and out of scope. Our focus in this paper is on the basic Federated Learning framework; how to adapt and evaluate it specifically for the task of mobile packet classification (as opposed to the image and text classification that is most commonly used for). The aforementioned attack scenarios in case of federated packet classification are deferred for future work. 


\section{Methodology}\label{sec:methodology}

\subsection{Problem Setup}

\begin{figure*}[!ht]
	\centering
	\includegraphics[width=0.99\textwidth]{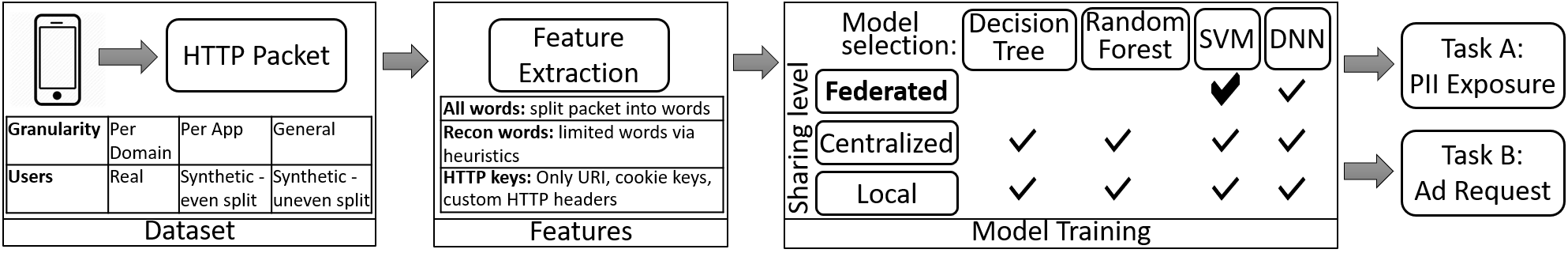}
	\caption{Overview of our pipeline for mobile packet classification. During {\em  training}, the input is  a packet trace with HTTP packets sent from mobile apps to remote destinations, labeled for the Task of interest (A: PII Exposure or B: Ad Request); the output is a per-app or a general machine learning model, which is trained in a local, centralized or federated way. During {\em testing}, the input is an HTTP packet, and the output is a binary label (indicating the presence or absence of PII or Ad request).} 
	\label{fig:overview}
\end{figure*}

	\textbf{Goal.} 
	We aim to train classifiers that use features extracted from HTTP requests coming out of mobile devices, to predict whether those packets contain information of interest, such as a PII exposure or Ad request.\footnote{Being able to classify packets  enables further action such as blocking the packet or obfuscating sensitive information, which is however, out of scope for this paper.}
	In order to train such classifiers, we need training data, \ie packet traces and labels indicating whether a packet contains the information of interest. We assume that training data are crowdsourced, \ie obtained and labeled on mobile devices and sent to a server that aggregates them and trains classifiers\footnote{This approach allows to train on characteristics of real users, as it was the case \eg in \cite{recon15}. Alternatively, training data can be obtained by automatically testing mobile apps \cite{haystack_paper, nomoads, antshield-arxiv}, which may allow to explore app behavior more systematically. Both approaches are valuable, but in this paper, we focus on the former.}. We also assume that the devices do not trust the server or other devices but they do want to contribute to the training and use the resulting global classifier. Our goal is to provide a methodology that enables devices to collaborate in training global classifiers, while avoiding to upload raw training data or even sensitive features to the server.


{\bf Federated Learning Approach.} To achieve this goal, we apply the Federated Learning framework (depicted on Fig. 1(c) and described in Section \ref{sec:related}) for the first time to the problem of mobile packet classification. This requires addressing several challenges and making design choices and optimizations, specific to our context, such as the following.

%
%

{\em Q1.} What packet features can best predict the labels of interest (\ie PII exposure or Ad request) in a packet?  Section \ref{sec:features} discusses how we 
 select \httpkeys features from HTTP packets, to achieve high classification performance while also meeting privacy and other constraints.
 
{\em Q2.} What model should we train with Federated Learning? Section \ref{sec:federated} compares different alternatives and proposes a \federated \svm framework.

{\em Q3.} What datasets should be used to train and test those classifiers?  Our training dataset consists of HTTP packets sent by mobile devices, labeled appropriately for each prediction task, \ie with binary labels to indicate PII exposure or Ad requests in each packet. Collection can be done using one of the existing VPN-based tools for intercepting traffic on the mobile device \cite{haystack-webpage, antmonitor15}, and labeling can be done using DPI \cite{recon15, antshield-arxiv, spawc}, blacklists \cite{easylists} or other tools \cite{nomoads}. Section \ref{sec:dataset_description} describes three real-world datasets we used in this paper. 

\textbf{Scope.} Our focus in this paper is on adapting and evaluating the Federated Learning framework specifically for mobile packet classification. An overview of our pipeline is provided on Fig. \ref{fig:overview}, and the rest of the section describes the details.

	The following considerations are out of the scope of this paper and deferred to future work. First, as discussed in Section \ref{sec:related}, there are known privacy attacks to Federated Learning, and proposed solutions (\eg secure aggregation and differential privacy) that can be added on the basic framework and are not considered in this paper. Second, our classifiers are trained on features extracted from HTTP or decrypted HTTPS traffic. This is a reasonable assumption today that (1) large portion of traffic is still unencrypted and (ii) several VPN tools \cite{haystack-webpage, antmonitor15} are able to successfully ``man-in-the-middle'' and decrypt HTTPS traffic {\em on the mobile device itself}, which is reasonable for the user to trust. As traffic gets increasingly encrypted and certificate pinning is adopted by more Android apps, intercepting HTTPS traffic will be more difficult\footnote{However, there are efforts from big companies, \eg  Facebook, to enable Developers/Researchers to proxy their own (or generated with testing accounts) traffic on Android devices \cite{fb_proxy}.} and classifiers will have access only to network level features (\eg IP/TCP headers, SNI, etc) rather than HTTP headers and payload. Federated Learning applies to that feature space as well, but this is out of scope for this paper.

\subsection{HTTP Features} \label{sec:features}

\begin{figure}[!ht]
	\centering
	\includegraphics[width=0.99\columnwidth]{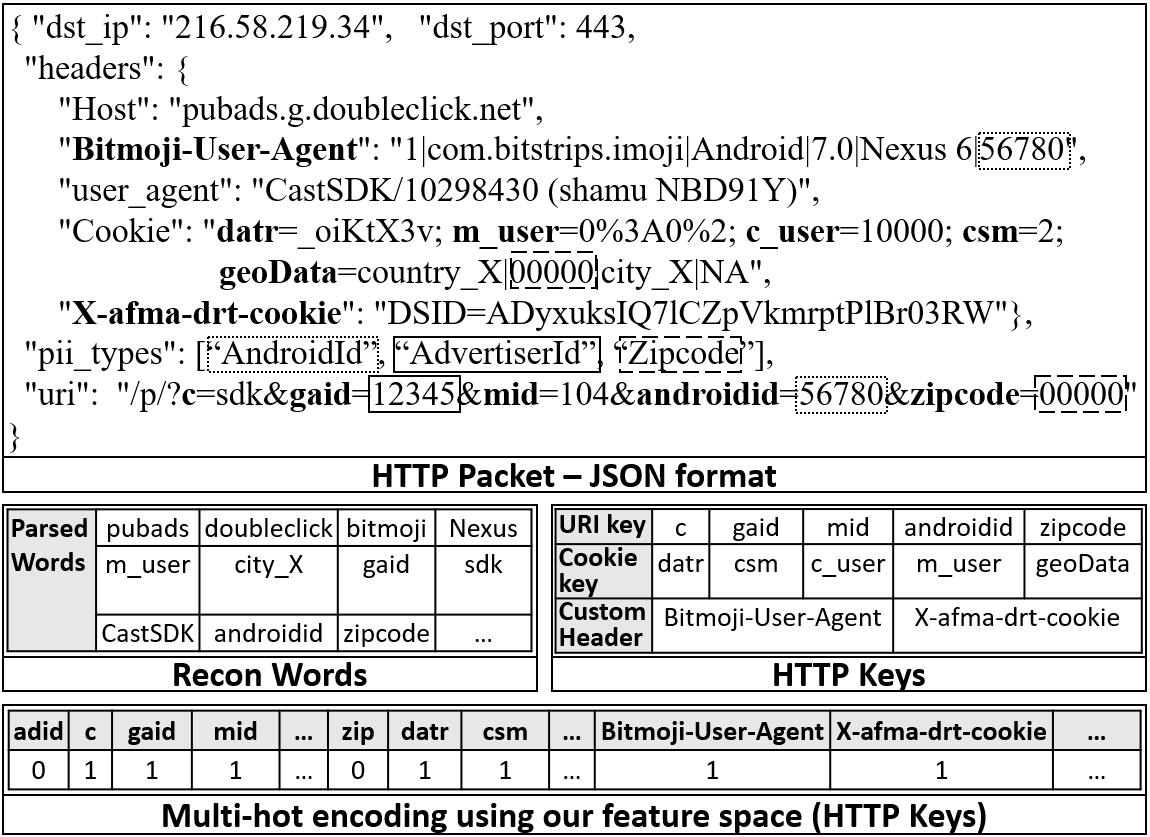}
	\caption{Example of an HTTP packet  in JSON format, where Android Id, Advertiser Id and zip code are sent by Bitmoji app  to an ad server (doubleclick.net). This packet would be labeled positive both for PII exposure and for Ad request. Our {\bf \httpkeys features} are highlighted in bold: these keys are defined by the HTTP protocol and extracted from (1) the \uri query keys (2) the \cookie keys and (3) custom HTTP headers (\eg ``Bitmoji-User-Agent''').  
	Compared to baselines (\allwords, \reconstyle), \httpkeys do not use sensitive information such as ``city\_X''.}
	\label{fig:json_example}
\end{figure}

{\bf Feature extraction.} We build on the approach introduced by \recon  \cite{recon15} and  used in follow-up classifiers \cite{antshield-arxiv, spawc, nomoads}: every HTTP packet is split into words  by using delimiter characters; the resulting words include  keys and values from all HTTP packet headers. Fig. \ref{fig:json_example} shows an example HTTP packet from Bitmoji (a mobile app that creates personal avatars), which sends several identifiers (Android Id, Advertiser Id and zip code) to an ad server. 
The question is which of these words to select as features 
to predict the presence of PII or Ad request in a packet, while facilitating Federated Learning, preserving privacy and meeting other constrains.  

There are several challenges when defining this feature space. First, we need to consider the trade-off between privacy and classification performance. This implies that we may not use some words that can help accurately classify packets, if these features themselves expose sensitive information (\eg part of URLs and domains can contain sensitive information about user's political views, medical conditions or sexual orientation); to that end we  do not use any values as features, only keys. Second, the feature space must have a small size (for high training speed, low memory and computation overhead for updates) and be fixed and known to all participating devices in the Federated Learning. 
%
Taking these constraints into account, we consider  three different feature spaces, two baselines and our proposed one:

\textbf{Baselines: \allwords vs. \reconstyle \cite{recon15, antshield-arxiv, spawc, nomoads}.}  Instead of considering the union of all words as the feature space (\allwords), \recon applied heuristics to remove the words that appear rarely and the most frequent words (stopwords,  
 which correspond to standard HTTP headers, common values such as values parsed from the user\_agent header). This results in removing some but not all values from consideration. 
 In particular, \recon discards the values after the ``='' delimiter, however certain values that do not follow this syntax will not be removed from the feature space and those might contain sensitive information. We refer to the remaining features as \reconstyle. The URI path also contains potentially sensitive information and words from URI path are also included in \reconstyle. 
Fig. \ref{fig:json_example} shows an example HTTP packet and a subset of the vocabulary of selected as \reconstyle, which includes some sensitive values such as ``city\_X''. 

\textbf{Our feature space: \httpkeys.}  During Federated Learning, all mobile devices and the server need to agree on the model and features they use, and they exchange information about model parameters. Both the features themselves and the parameter updates can potentially contain sensitive information. To avoid that privacy risk, we  purposely limit our feature space to use only non-sensitive keys from HTTP packets. 
More specifically, we consider the structure of HTTP packets and extract features from: (1) the \uri query keys,  (2) the \cookie keys, and (3) custom HTTP headers; and (4) whether or not there is  a file request in the packet. We refer to the collection of these features as \httpkeys. 

First, consider the \uri: it typically contains a relative path on a given domain and queries, usually built using key-value pairs separated by ``\&'' characters. Sensitive information in the \uri typically appears in relative paths and query values, while \querykeys represent API calls to the destination domain. We only use \querykeys as features.  We do not extract any features from the domain and the \uri path, since it may contain sensitive information about the user. 
Second, we include keys from the \cookie field. 
Query keys from these two fields are sufficient to extract features for most packets in our datasets.

Third,  to differentiate more packets, we extract 
custom HTTP headers, which are defined by apps and can embed sensitive information about users. In recent years, apps have started using custom headers 
in order to provide app specific functionality. 
We remove the standard 
HTTP headers \cite{http_header_list} from all HTTP headers in order to retrieve the custom ones.

Fourth, if a packet does not have any keys in the \uri field, \cookie header or custom HTTP headers, we include \filerequest  -- a boolean feature that indicates the presence of a file request. In most cases this will be a benign activity such as requesting static HTML content. 
Packets in the datasets that do not contain any of the four features, which we refer to as \keyless, are excluded from our  pipeline. 

{\bf Feature Space Size.}  Considering \httpkeys as features already reduces the feature space. However, the feature space size varies widely across apps and users. Different apps have different APIs (which may lead to different \querykeys and custom HTTP headers)  and they may contact different destination domains. We differentiate between two broad categories of apps according to the number of contacted domains: apps with or without  \webview. \webview apps can contact any domain and present web content in the \webview; examples include browsers or social media apps like Facebook. Apps without \webview are more likely to only contact a small fixed set of domains, such as backend servers, analytic and advertisement services.  Apps with \webview present  new challenges, since the feature space could explode with hundreds of features from every new user, who visits previously unseen domains. We discuss more about \webview apps and their impact on the feature space size in Section \ref{sec:dataset_description}.


\textbf{Vocabularies.} Vocabularies are used in machine learning models with non-numerical features; in our case the vocabulary is the set of unique words in the dataset. Throughout this paper, we refer to vocabulary and feature space interchangeably. 
In this paper, we use Multi-hot encoding to represent the extracted words per packet. 
A Multi-hot encoding is a sparse binary vector with the length of the vocabulary such that it is has 1s at the locations of words in the vector; 0 otherwise. 
 An example is shown at the bottom of Fig. \ref{fig:json_example}. 
We  use the same feature space for both classification tasks (Ad request and PII exposure detection), because there is a relation between the two tasks: apps use PII information for serving ads.

In Federated Learning,  the vocabulary must be fixed and shared a-priori between all mobile devices and the server. \reconstyle potentially expose sensitive information during the construction of the shared feature space.
Fixing a vocabulary across multiple users is successful when the feature space is fixed \ie for apps without \webview. The intuition is that a single user might not explore the entire API of a service, but across multiple users this is more likely to happen. 




\subsection{Model Selection: Federated SVM} \label{sec:federated}

 Once the feature space is fixed, our goal is to train a classifier using Federated Learning. The first step is to select the classification model, \eg  Decision Trees (DTs), Random Forest (RF), Deep Neural Networks (DNNs), Support Vector Machines (\svms), etc.  The next step is to train that model in a \federated way (Fig. \ref{fig:distributed_learning_figure_v2}(c)) and compare it to its \centralized  (Fig. \ref{fig:distributed_learning_figure_v2}(b)) and \standalone versions (Fig. \ref{fig:distributed_learning_figure_v2}(b)).  The choices we evaluate across these two dimensions (\ie classification model and degree of collaboration among users) are summarized under ``Model Training'' on  Fig. \ref{fig:overview}.

{\bf Selecting \svms as the Classification Model.} State-of-the-art classifiers for mobile packets have trained DTs \cite{recon15,spawc,nomoads} 
to predict PII exposure or Ad requests based on features extracted from outgoing HTTP packets. DTs were chosen primarily because of their interpretability for small tree sizes and their efficiency for on-device prediction \cite{antshield-arxiv, spawc, nomoads} -- most packets are classified after a few levels in the tree. Unfortunately, DTs do not naturally lend themselves to federation, since there is no framework for ``aggregating'' multiple decision trees collected from multiple devices at the server.  In this paper, we use DTs as baseline centralized models.

Federated averaging was developed for models based on Stochastic Gradient Descent (SGD), primarily DNNs \cite{original_federated, mcmahan2017federated}. In SGD-based models, the mobiles and the server exchange gradient updates, 
 and the server simply  averages the local gradients to update the global model. Unfortunately,  DNNs require a large number of  samples to train, which is costly (in device resources and user experience) to obtain and train on mobile devices.

While Federated Learning is most commonly used to train DNNs, it applies to any SGD-based model that lends itself to aggregation at the server.
In this paper, we select \svms. Compared to DTs: \svms are SGD-based  -- thus amenable to federation,  achieve similar \fscore (due to the simple binary vector representation that comes from the multi-hot encoding) and interpretability (through weight coefficients).  Compared to DNNs:  (1) \svms use fewer parameters which means less computation, communication and faster training; (2) Linear Kernel \svms have convex loss functions where more principled guarantees can be provided for convergence during training; (3) \svms usually perform better than DNNs on datasets with limited size; (4) \svms are easier to interpret. 


%

\begin{algorithm}[t]
	\small
\SetAlgoNoEnd
\SetAlgoNoLine
	Given $K$ clients (indexed by $k$); $B$ local minibatch size; $E$ number of local epochs; $R$ number of global rounds; $C$ fraction of clients and $\eta$ learning rate.\\
	{\bf Server executes:}\\
	\SP Initialize $w_0$\\
	\SP \For{each round t = 1,2, ... $R$} {
	\SP 	$m \leftarrow max(C\cdot K,1)$\\
	\SP 	$S_t \leftarrow$ (random set of $m$ clients)\\
	\SP 	\For{each client $k \in S_t$ in parallel} {
	\SP 		$w_{t+1}^k \leftarrow ClientUpdate(k,w_t)$\\
	\SP 	}
	\vspace{-8pt}
	\SP 	$w_{t+1} \leftarrow \sum_{k=1}^K \frac{n_k}{n}w_{t+1}^k$\\
	\SP }
	\vspace{4pt}	
	{\bf ClientUpdate($k,w$):}\\
	\SP $B_k \leftarrow$ (split of local data into batches of size $B$)\\
    \SP  \For{each local epoch i from 1 to $E$}{%
	\SP 	\For{batch $b\in B_k$} {
    \SP $w \leftarrow w - \frac{\eta}{B} \sum_{i\in B_k} y_i \cdot x_i$,  when $y_i  (w_i x_i)<1$\\
    	}
	}
	\SP  \Return{$w$ to server}
\caption{Federated SVM}
\end{algorithm}

{\bf Federated \svm.} In this paper, we use  {\em Federated \svm} with linear kernels as the core of our federated packet classification. 
Linear kernel \svm minimizes the following objective function, $f$, over weight vector $w$: 
\begin{equation}
f(w,X,Y)= \sum_i l(w,x_i,y_i) + \lambda ||w||^2,
\vspace{-5pt}
\end{equation}
where $x_i$ is the feature vector (\ie the Multi-hot encoding for a packet), $y_i$ is the binary label and the loss function is called the Hinge loss:
$l(w,x,y) =max(0, 1- y \cdot (w\cdot x))$.
%
Pegasos previously  applied the SGD algorithm for \svm \cite{Shalev-Shwartz2011}. 
The Hinge loss function is convex and has the necessary sub-gradients, \ie if $y\cdot w\cdot x < 1$, then $\triangledown l(w,x,y)=  -y \cdot x$, otherwise $0$. This step is easily added to the SGD algorithm, but more importantly to Federated Averaging \cite{McMahan2016}. 

Algorithm 1 shows our algorithm for Federated  SVM: we apply the SVM-based 
 gradient updates to the Federated Averaging algorithm from \cite{McMahan2016}.
Federated \svm trains an \svm model distributively over $K$ clients (corresponding to mobile devices), where $C$ fraction of the clients update their model in each round and all clients update the global model by averaging their model parameters. A client update consists of multiple local epochs, $E$, and minibatch split of local data into $B$ batches similar to standard SGD algorithm. Clients compute the SGD update based on the above Hinge loss.

The Federated SGD algorithm is a special case of Federated Averaging for $C=1$, $E=1$, $B=\infty$ \cite{McMahan2016} (\ie use every client in a round with a single pass on all their local data once). Usually, we look to push more computation to the clients by setting $E>1$ and  $B$ to a small number, and use a small fraction of clients $C$ in each global round.  \cite{McMahan2016} explores the trade-off between these hyper-parameters and shows how to decrease the global number of rounds $R$ required to reach a target accuracy on the test sets for image classification and next word prediction. The Federated Learning framework trains a shared model, hence the feature space has to be fixed and shared across multiple users. 
 The size of feature space affects parameter updates, and thus communication costs during model training.  




{\bf \federated vs. \centralized and \standalone models.}
Once we have fixed the feature space and underlying model (\svm with linear kernel), we compare the 
\federated vs. \centralized and \standalone 
models, as shown in overview depicted 
in  Fig. \ref{fig:distributed_learning_figure_v2}. 
\begin{itemize}
	\item \standalone models are trained on data available  on each device, similar to previous works \cite{ recon15, antshield-arxiv, spawc, nomoads}. Nothing is shared outside the devices, thus preserving 
	 privacy but not classification performance.
	\item \centralized models: devices upload their training data to a server, and a global model is trained at the  central server, similar to previous works of \antshield \cite{spawc, antshield-arxiv}, \recon \cite{recon15} and \nomoads \cite{nomoads}. This approach trains better classifiers but shares potentially sensitive training data.
	\item \federated models: devices do not share training data with the server, but send model parameter updates to the server, which then aggregates, updates the global model and pushes it to all devices; the process repeats until convergence. 
\end{itemize}



\begin{table*}[!ht]
	\setlength\tabcolsep{1pt} 
	\centering
	\small
\begin{tabular}{l|llc|c|cccc|cccc}
	\toprule
	&  &  &  & \multicolumn{1}{|c|}{\textbf{Prior Work}} & \multicolumn{4}{c|}{\textbf{\httpkeys}} &  & &  &  \\ \midrule

	\multirow{2}{*}{\textbf{Dataset}} & \multirow{2}{*}{\textbf{\#Apps/Users}} & \multirow{2}{*}{\textbf{\#Packets}} & \multirow{2}{*}{\textbf{\#Ads/PII}} & \textbf{\#Features} & \textbf{\#\uri}  & \textbf{\#\cookie} & \textbf{\#Custom} & \textbf{\#File} & \textbf{\#\Keyless} & \textbf{\#Destination} 
	 & \multirow{2}{*}{\textbf{Time}} \\
	& & & & 
	\textbf{All/\reconstyle} & \textbf{keys} &  \textbf{keys} & \textbf{Headers} & \textbf{Requests} & \textbf{POST Packets} & \textbf{Domains}  \\
	\nomoads & 50/(synthetic) & 15,351 & 4,866/4,427 & 12,511/6,743 & 2,580 & 216 & 204 & 3,342 & 2,334/2,123 & 366 & 2017\\
	\antshield & 297/(synthetic) & 41,757 & -/8,170 & 39,304/19,778 & 3,855 & 302 & 609 & 4,644 & 8,836/777 & 674 &  2017 \\
	\Inhouse & \multirow{2}{*}{1/8 real} & \multirow{2}{*}{84,716} & \multirow{2}{*}{-/3,424} & \multirow{2}{*}{40,936/22,714} & \multirow{2}{*}{7,573} & \multirow{2}{*}{3,591} & \multirow{2}{*}{47} & \multirow{2}{*}{12,903} & \multirow{2}{*}{13,786/153} & \multirow{2}{*}{1,607}  & \multirow{2}{*}{2015} \\
	Chrome & & & & & & & & & &  \\
	\Inhouse & \multirow{2}{*}{1/10 real} & \multirow{2}{*}{33,580} & \multirow{2}{*}{-/1,347} & \multirow{2}{*}{11,921/6,718} & \multirow{2}{*}{4,370} & \multirow{2}{*}{1,160} & \multirow{2}{*}{19} & \multirow{2}{*}{172} & \multirow{2}{*}{12/0} & \multirow{2}{*}{861}  & \multirow{2}{*}{2015} \\
	Facebook  & & & & & & & & & & \\
	\bottomrule
\end{tabular}

	\caption{ Summary of each dataset in terms of features (our \httpkeys method vs. \reconstyle vs. \allwords), 
	total amount of data, users, visited domains and supported classification tasks.} 
	\vspace{-4pt}
	\label{table:datasets_summary}
\end{table*}

\section{Datasets Description}\label{sec:dataset_description}


We use three real-world datasets, summarized on Table \ref{table:datasets_summary}, to evaluate the performance of our federated approach, w.r.t. two   packet classification tasks, \ie PII exposures and Ad requests.

\textbf{\nomoads dataset \cite{nomoads, nomoads_data}.} The \nomoads dataset is made publicly available at \cite{nomoads_data} by the authors of \cite{nomoads}. It consists of HTTP and unencrypted HTTPS packets, labeled with  ad requests and PII exposures they may contain, from 50 most popular apps in the Google Play Store. The data was generated via manual testing (interacting with each app for five minutes) with test accounts and there were no human subjects involved. The data was labeled by using state-of-the-art filterlists \cite{easylists} of ad serving domains initially and then if an ad was still appearing during manual testing, a rule was produced manually via visual inspection to detect ads in the last few packets. \nomoads is the only dataset that contains state-of-the-art labels for advertising.

\textbf{\antshield dataset \cite{spawc, antshield-arxiv, antshield_data}}. The \antshield dataset is made publicly available at \cite{antshield_data} by the authors of \cite{spawc, antshield-arxiv}. This dataset contains HTTP and HTTPS  packets labeled to indicate if they contain a PII exposure or not. W.r.t. PII it is similar to \nomoads dataset but it is richer since it contains data from more apps for the PII task. The data was generated with manual and automated testing. The manual tests included interacting for five minutes with the top 100 most popular apps (according to AppAnnie \cite{appannie}). 
The automated tests included top 400 apps and a monkey script \cite{monkey} was performing 1,000 random actions over ten minutes per app. 
  We combine the \antshield data generated from manual and automated testing to a single dataset, from which we consider the 297 apps out of 400, that generated HTTP or HTTPS traffic. 
We would like to note that there are other publicly available datasets with PII labels, \ie extended dataset with many apps and their different versions during 8 years \cite{ren2018bug}, however this dataset does not include logs from all HTTP fields packets and it would be impossible to extract our chosen features from URI and  Cookie fields and the presence of custom headers. Another publicly available dataset is \cite{recondata} with labeled HTTP packets. However, the mapping of packets to the app that generated it, is not reliable and since we sample the data per app in order to create synthetic users, we decided to use the \antshield dataset as an example dataset for PII task.

\textbf{\Inhouse Datasets with real users.} This is a dataset we collected \inhouse from 10 real users 
who contributed their 
packet traces 
for a period of 7 months\footnote{The study was approved by the University of California Irvine's Institutional Review Board (IRB).}. The packet traces were collected by running Antmonitor \cite{antmonitor-arxiv}: an open-source VPN-based tool that intercepts outgoing network traffic generated from each Android app. In order to run our machine learning algorithms, 
we have preprocessed the raw packet traces into JSON, by keeping only HTTP packet-level information. 
We redacted all user sensitive information with a prefix and the type of PII it contained (\eg prefix\_email) and labeled the packets with exposures if they contained one of these scrubbed PII exposures. To evaluate Federated Learning, 
 we consider the two most popular apps across all ten users, which are Chrome and Facebook with 8 and 10 users in total respectively. We consider each of these two apps as  a separate dataset: \inhouse Chrome and \inhouse Facebook. 

\textbf{Packet Classification Tasks.} In all three datasets, a packet is considered to have a {\em PII exposure}, if it contains  some personally identifiable information (PII), including the following: (i) device identifiers, such as  IMEI, Device ID, phone number, serial number, ICCID, MAC Address; (ii) user identifiers such as first/last name, Advertiser ID, email, phone number; (iii) Location: latitude and longitude coordinates, city, zip code.
This is a subset of PII as defined in prior work, but our framework can be used to detect additional PII types if the corresponding labeled ground truth is provided.
If a packet contains at least one of these sensitive fields,  we  assign label 1 to the packet, otherwise 0. 
For the ad prediction task, if a packet contains an {\em Ad Request} it is labeled as 1, otherwise 0. 
These labels are 
available for each packet in the datasets considered.


{\bf Summary of the Datasets.} Table \ref{table:datasets_summary} summarizes the feature space, as relevant to our federated learning framework, including: total number of unique features (\uri keys, \cookie keys and custom HTTP headers), total number of packets, total \keyless packets and how many of them were POST requests, total packets that contain a \filerequest only but no other feature, 
and unique destination domains.
 We do not include  HTTP POST packets in our training or testing, for \keyless packets, \ie packets without any features (\querykeys in the \uri or \cookie field, custom HTTP headers, or \filerequest). There is no standardized syntax for the POST body in order to obtain only the keys without parsing the values too. Therefore, for privacy reasons we decided to not parse them at all and to discard such packets from our experiments.

The \antshield dataset contains the most apps and packets with a PII exposure (8,170), while \inhouse Chrome contains the most packets (84,716) and the highest number of unique domains (1,607).  In  the \nomoads dataset, the feature space 
has 12,511 features with \allwords from the HTTP packet (including values) and 6,743 using \reconstyle. On the other hand, \httpkeys  uses only 3,001 features (Table \ref{table:datasets_summary}: sum of  \uri, \cookie keys, custom headers + 1 for \filerequest), which is less than half of the \reconstyle. Similarly, in the \antshield dataset the feature space increases from 4,767 with \httpkeys, to 19,778  \reconstyle and to 39,304 with \allwords. This explosion of feature space affects the training speed, the size of the trained models and might expose sensitive information of user data (\ie values to sensitive keys). The benefit of our \httpkeys approach is the following: (1) our significantly reduced feature space can describe both prediction tasks (Ads and PII), (2) users share limited sensitive information, without sacrificing classification accuracy and (3) the reduced number of features leads to smaller models and faster training, which is important in mobile environments.

\begin{figure}[!t]
	\centering
	\includegraphics[width=0.90\columnwidth]{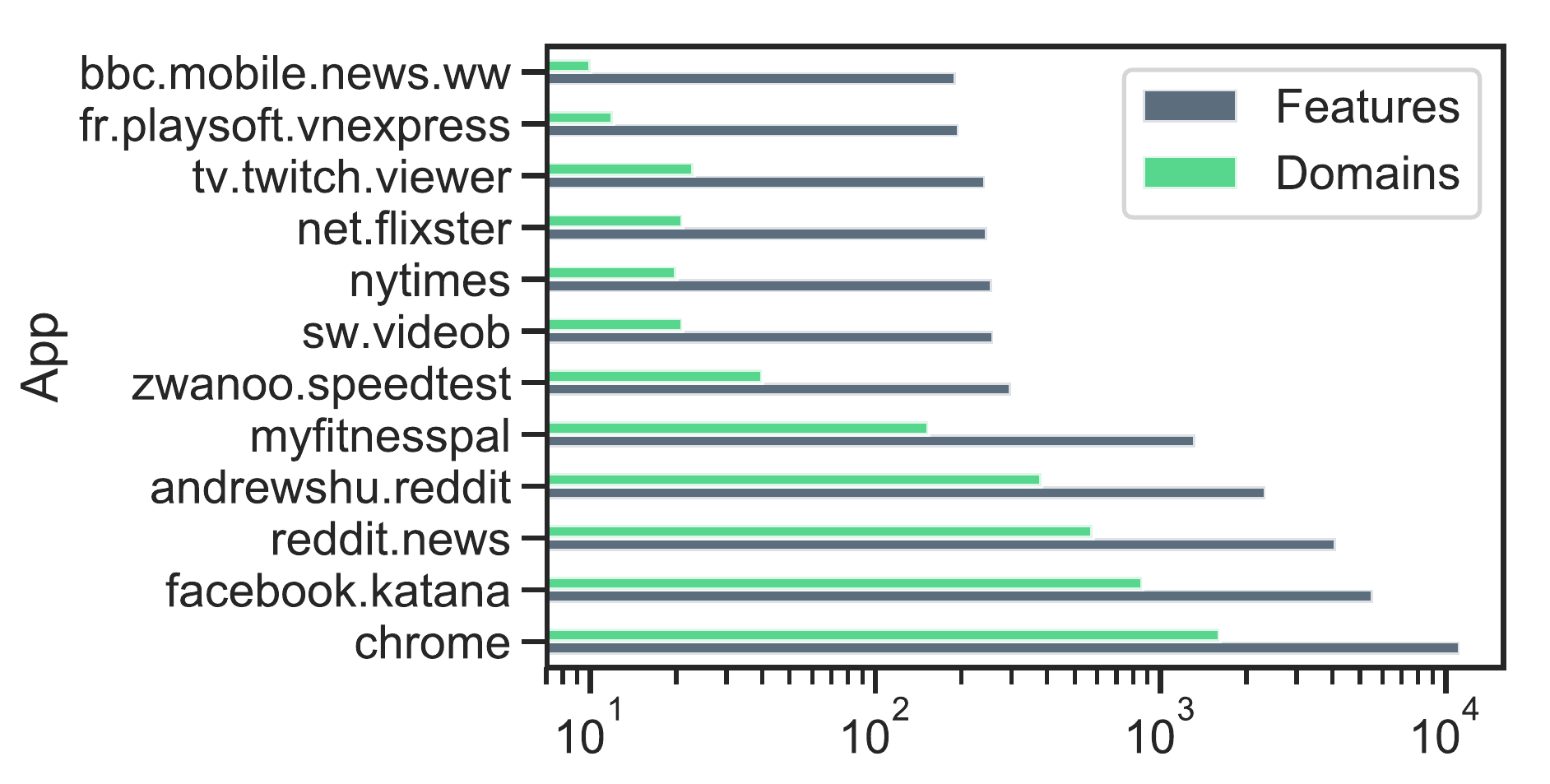}
	\caption{ 
		Number of features and domains for the top 12 apps with most features from our in-house dataset. The number of features correlates with number of visited domains.}
	\label{fig:features_vs_domains_minas_top}
\end{figure}

\textbf{\webview vs. non-\webview apps.} Figure. \ref{fig:features_vs_domains_minas_top} shows the distribution of features and domains for the top 12 apps with most features from our \inhouse dataset. There is a positive correlation between the number of features and visited domains for each app. 
\webview apps, such as Facebook and Chrome, have the most features, as expected. 
Table \ref{webview-summary} shows the feature space for Chrome and Facebook 
in all our datasets. 
The feature space of  \webview apps depends on the usage of each app, \eg the duration (in terms of packets), user behavior (in terms of domains visited). 
We observe the explosion of the feature space in our \inhouse dataset, where Chrome has only 370 shared features across 8 users, but the union has 11,212. 
Similarly, the Facebook 
app has only 14 common out of 5,550 features  across 10 users. 
%
In contrast, non-\webview apps have smaller feature spaces due to their limited number of contacted domains. In this paper, we assume that the datasets contain all possible visited domains and the feature set is fixed.

\begin{table}[!t]
	\setlength\tabcolsep{5pt} 
	\small
	\centering
	\begin{tabular}{c|ccll}
		\toprule
		\multirow{2}{*}{\textbf{Chrome}} & Intersection & Union & \#Packets & \#Domains  \\
	
		&  features & features & & \\		\midrule		
		\Inhouse & 370 & 11,212 & 84,716 & 1,607  \\
		\antshield & - & 75 & 206 & 15  \\
		\nomoads & - & - & - & -  \\ \bottomrule
		\multirow{2}{*}{\textbf{Facebook}} & \textbf{Intersection} & \textbf{Union} & \textbf{\#Packets} & \textbf{\#Domains}  \\ 
		&  \textbf{features} & \textbf{features} & & \\		\midrule
		\Inhouse  & 14 & 5,550 & 33,580 & 861 \\
		\antshield  & - & 63 & 110 & 4 \\
		\nomoads & - & 820 & 392 & 82 \\ \bottomrule
	\end{tabular}
	\caption{Two \webview apps and comparison of their feature space 
		in our datasets. 
		We present the intersection/union of features, number of packets and domains across all datasets.} 
	\label{webview-summary}
\end{table}


\section{Results}\label{sec:results}


\textbf{General Setup.}  Table \ref{all_options} lists the possible options for evaluation based on our pipeline defined in Section \ref{sec:methodology}.  We compare the \federated model to \standalone and \centralized models where the test data comes either from a user or is the union of test data from all users. 
We train only general and per-app  models, 
but no per-domain model (as there can be too many domains and it would be impractical to train a separate model for each). In each scenario, we describe the 
evaluation setup, rationale and results in terms of classification accuracy, communication and computation cost.



To evaluate the classification performance, we split the available data into 80\% train and 20\% test data in order to compute \fscore  \cite{f1_score_paper} on the positive class (\ie Ad request or PII exposure is detected). Furthermore, in each of the experiment scenarios, we apply standard machine learning techniques to train efficient classifiers. Before training, we balance our dataset so that it contains an even amount of positive and negative examples to avoid training with a bias towards the most frequent class. For each of the following experiments we train and test five times each model (unless otherwise mentioned) to obtain an average \fscore.


\begin{table}[!t]
	\centering
	\small
	\setlength\tabcolsep{3pt}
	\begin{tabular}{l|l|l|l}
		\toprule
		Pipeline& \multicolumn{3}{c}{Options} \\ \midrule
		\multirow{2}{*}{\textbf{Dataset:}} & \multirow{2}{*}{\nomoads} & \multirow{2}{*}{\antshield} & \Inhouse Chrome \\
		& & & Facebook \\ \hline
		\multirow{2}{*}{\textbf{Users:}} & \multirow{2}{*}{Real users} & \Even split  &  \Uneven split \\
 		& & with k users &	with k users	\\ \hline
		
		\textbf{Classifier} & \multirow{2}{*}{(Per domain)} & \multirow{2}{*}{Per App} & \multirow{2}{*}{General}\\
		\textbf{Granularity:} & & &  \\ \hline
		\multirow{2}{*}{\textbf{Models:}}&  \multirow{2}{*}{\federated \svm}  &  \multirow{2}{*}{\standalone \svm} & \centralized \svm  \\
		& & &  /baselines \\ \hline
		\textbf{Tasks:} & PII exposure & Ad request &  \\ 
		\bottomrule
	\end{tabular}%
	\caption{Parameters of the 
	Evaluation Setup.}
	\label{all_options}
\end{table}

\textbf{Creating synthetic users.} \nomoads and \antshield datasets do not come from real users, since they were produced manually by their corresponding authors or automatically via running monkey scripts with random actions. 
We create synthetic users by sampling from the available data in order to test our Federated approach for packet classification.
We developed two different approaches to partition the data into $k$ synthetic users: a random split into equal amounts of data (\even split) and a random split of data with random sizes of sampled data so that each user contains a different amount of packets (\uneven split). 
For the \uneven split, we used a minimum threshold, 30\%, of available app data to be assigned to each user. 
We test both methods and show their results, since the advantage of Federated Learning is that it can handle various distributions of data across participating users. 

For both synthetic and real users, we apply the train and test split per user to train \standalone, \centralized or \federated classifiers. In addition, we show in Section \ref{sec:inhouse}, that training on a subset of users can provide good classifiers for all users.



\subsection{Scenario 1: Centralized Models} 

\textbf{Setup 1.} In this experiment, we use the following setup from Table \ref{all_options}:  \textit{Dataset:} \nomoads. \textit{Users:} None. 
\textit{Classifier Granularity:} General. 
 \textit{Models:} \centralized \svm (linear and non-linear kernel, SGD) and baselines (DT, RF). \textit{Tasks:} PII exposure and Ad request. 
 The goal  is to validate our choice of \federated model (\svm with SGD) and feature space (\httpkeys and \filerequest) in the rest of the paper.


\begin{figure}[t!]
	\centering
	\includegraphics[width=0.80\columnwidth]{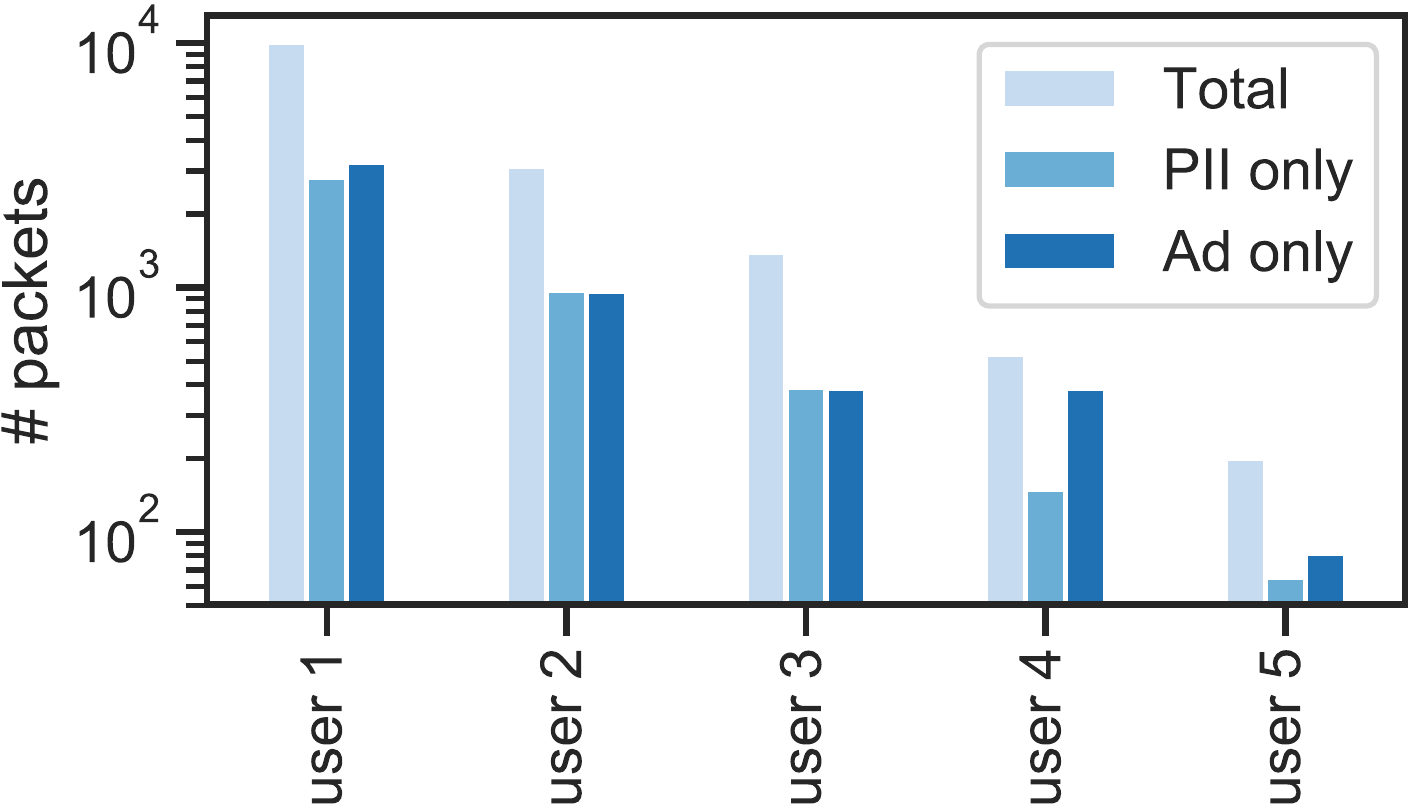}
	\caption{ 
		\uneven split of \nomoads across 5 synthetic users.}
	\label{fig:nomoads_random_partition_5_users_distribution}
\end{figure}

\begin{table*}[!ht]
	\setlength\tabcolsep{4pt} 
	\centering
	\small
	\begin{tabular}{ccc|cc|cc|cc}
		\toprule
		\textbf{Feature Space (\# Features)} & \multicolumn{2}{c}{\httpkeys (3000)} & \multicolumn{2}{c}{\httpkeys + \filerequest (3001)} & \multicolumn{2}{c}{\reconstyle (6,580)} & \multicolumn{2}{c}{\allwords (12,195)} \\
		\textbf{Task} & Ads & PII & Ads & PII & Ads & PII & Ads & PII \\
		\textbf{Centralized Classifier} & \textbf{\fscore} & \textbf{\fscore} & \textbf{\fscore} & \textbf{\fscore} & \textbf{\fscore} & \textbf{\fscore} & \textbf{\fscore} & \textbf{\fscore} \\ \midrule
		Decision Tree (DT) & 0.936 & 0.98 & 0.854 & 0.95 & 0.98 & 0.984 & 0.979 & 0.983 \\
		Random Forest (RF) & 0.938 & 0.981 & 0.861 & 0.949 & 0.982 & 0.986 & 0.979 & 0.987 \\
		SVM with SGD & 0.929 & 0.975 & 0.838 & 0.944 & 0.971 & 0.981 & 0.975 & 0.979 \\
		SVM linear kernel & 0.933 & 0.979 & 0.857 & 0.952 & 0.984 & 0.984 & 0.981 & 0.984 \\
		SVM rbf kernel & 0.706 & 0.762 & 0.625 & 0.744 & 0.785 & 0.756 & 0.761 & 0.719 \\ \bottomrule
	\end{tabular}
	\caption{{\bf Results 1a and 1b.} The performance of various ML models on the \nomoads dataset for the two tasks: Ads and PII prediction. The reported \fscore is averaged, after training and testing each model  5 times. We show that \svm with SGD performs as well as DT and RF. 
		We increase the feature space (packet information used) from left to right. \httpkeys results in significant reduction in the number of features, while achieving high \fscore for PII (0.94) and for Ads prediction (0.85). 
	}
	\label{models}
\end{table*}

\textbf{Results 1a: \httpkeys vs. \reconstyle features.} 
 In Table \ref{models}, we compare various \centralized classifiers on four different feature spaces:
\httpkeys (3,000 features), \httpkeys with \filerequest, \reconstyle (6,580 features), \allwords (12,195 features).
\httpkeys with \filerequest uses a smaller feature space (3,001 features) but achieves an \fscore above 0.94 and 0.85 for PII and Ads, respectively.

Adding the \filerequest feature includes more packets which results in a classification loss of approximately 8\% and 3\% for Ads and PII prediction accordingly. The drop in performance is slightly larger in case of Ad predictions, since our feature space does not include information about domains 
that is important for 
this task as shown in \cite{nomoads}. Prior work \cite{recon15, antshield-arxiv, spawc, nomoads} uses domain information in addition to other potentially sensitive features, and achieves higher \fscore. There is always a trade-off between privacy and utility, however, our defined feature space and the distributed framework are good steps towards private packet classification, without significant loss in classification performance.

\textbf{Results 1b: \svm with SGD performs similarly to Decision Trees.} We compare \svm with SGD to state-of-the-art baseline models, such as Decision Trees and Random Forest (used in \antshield \cite{antshield-arxiv}, \recon \cite{recon15}, \nomoads \cite{nomoads}). Table \ref{models} compares their performance on the \nomoads dataset. 
For all feature spaces (i)  the linear \svm and \svm with SGD   perform similarly to Decision Tree and Random Forest; 
and (ii) \svm with a non-linear kernel (rbf) seems to not generalize well and it is likely to overfit to data. 
Therefore, we select  \svm with SGD as the basis of our Federated Learning framework.

\subsection{Scenario 2: NoMoAds for PII, Ad Request}

\textbf{Setup 2a.} We use the following setup from Table \ref{all_options}. \textit{Dataset:} \nomoads.\textit{Users:} \Even and \Uneven split across 5 synthetic users; the distribution of data for \uneven split is depicted  on Fig.\ref{fig:nomoads_random_partition_5_users_distribution}.  
 \textit{Classifier Granularity:} General. \textit{Models:} \federated \svm vs. \centralized \svm. \textit{Tasks:} PII exposure and Ad request. We set local epochs to $E = 5$, batch size to $B = 10$ and we use all users by setting the fraction $C = 1.0$, as we use only 5 synthetic users due to the limited size of the dataset. 

\begin{table}[!t]
	\small
	\centering
	\begin{tabular}{llllll}
		\toprule
		\setlength\tabcolsep{1pt} 
		 \multirow{2}{*}{}&  \multirow{2}{*}{} & \multicolumn{2}{l}{ \textbf{\Uneven split}} & \multicolumn{2}{l}{\textbf{\Even split}} \\
		 & & \multicolumn{2}{l}{\textbf{\fscore}} &   \multicolumn{2}{l}{\textbf{\fscore}}\\
		\textbf{Trained on} & \textbf{Tested on} & \textbf{Ads} & \textbf{PII} & \textbf{Ads} & \textbf{PII}\\ \midrule
		\federated & user 0 test & 0.83 & 0.96 & 0.84 & 0.95 \\
		\federated & user 1 test & 0.92 & 0.96 & 0.81 & 0.95 \\
		\federated & user 2 test & 0.86 & 0.95 & 0.84 & 0.95 \\
		\federated & user 3 test & 0.63 & 0.97 & 0.88 & 0.92\\
		\federated & user 4 test & 0.85 & 0.96 & 0.86 & 0.96 \\
		\federated & all test data & 0.84 & 0.96 & 0.85 & 0.95 \\
		\midrule
		\standalone user 0 & user 0 test & 0.82 & 0.95 & 0.78 & 0.9 \\
		\standalone user 1 & user 1 test & 0.89 & 0.94 & 0.8 & 0.92 \\
		\standalone user 2 &user 2 test & 0.8 & 0.9 & 0.79 & 0.93  \\
		\standalone user 3 & user 3 test & 0.64 & 0.82 & 0.83 & 0.9  \\
		\standalone user 4 & user 4 test & 0.77 & 0.87 & 0.81 & 0.91 \\
		\centralized  & all test data & 0.85 & 0.96 & 0.84 & 0.94  \\ \bottomrule
	\end{tabular}
	\caption{ 
	{\bf Results 2a.} \federated performs as well as  \centralized training and significantly outperforms \standalone models. We show the \fscore for each user, when testing on (i) their hold-out test data and on (ii) the union of all users test data.}
	\label{nomoads_pii_federated_vs_centralized}
\end{table}

\begin{table}[!t]
	\setlength\tabcolsep{3pt} 
	\centering
	\small
	\begin{tabular}{lllll}
		\toprule
		\textbf{} & \multicolumn{2}{l}{------- \Uneven split -------} & \multicolumn{2}{l}{------ \Even split ------} \\
		$C$ & $B = \infty$ & $B = 10$ & $B = \infty$ & $B = 10$ \\ \midrule
		\multicolumn{5}{l}{\textbf{Task: Ads with target \fscore= 0.85, $E = 1$}} \\ \midrule
		0.05 & 36.6 [24, 58]  & 22.4 [11, 29]  & 25 [19, 30]    &  33.4 [25, 43]\\
		0.1 & 15 [10, 20] &  15.2 [9, 24] & 14 [11, 23] & 23 [13, 34]  \\
		0.2 & 10 [8, 13]  &  6.8 [5, 10] & 8.6 [7, 14]  & 11 [6, 17] \\
		0.5 & 2.6 [2, 4]  &  3 [2, 4] & 4.4 [3, 6] & 8 [6, 11]  \\
		1.0 & 1.6  [1, 2] & 2.4 [1, 4] & 2.6 [2, 4]  &  4.8 [4, 6]\\ \midrule
		\multicolumn{5}{l}{\textbf{Task: Ads with target \fscore= 0.85, $E = 5$}} \\ \midrule
		0.05 & 43.6 [40, 48]  & 49 [27, 75]  & 34.8 [27, 53]   & 48.8 [43, 63]  \\
		0.1 &  21.2 [13, 28] &  20.8 [17, 26] & 22.6 [19, 27] &  22.4 [18, 27]\\
		0.2 & 12 [8, 15]  &  10 [7, 11] & 9.2 [8, 12] & 10.6 [10, 12]  \\
		0.5 & 3.4 [2, 6]  & 4.2 [3, 5] & 3.8 [3, 5]  & 5.6 [3, 11]  \\
		1.0 & 1 [1, 1]  & 1.2 [1, 2]  & 2.8 [2, 6]  & 3.4 [2, 7]  \\ \midrule
		\multicolumn{5}{l}{\textbf{Task: PII with target \fscore= 0.95, $E = 1$}} \\ \midrule
		0.05 & 30 [19, 37]] & 28.8 [21, 40]   & 27.8 [21, 33]  &  27.8 [23, 31]\\
		0.1 & 14.2 [9, 18] & 15.6 [12, 18]  & 16.4 [13, 19] & 16.8 [16, 18] \\
		0.2 & 7.4 [4, 9] & 7.4 [5, 12]   & 7.4 [6, 9] & 7.2 [6, 10]  \\
		0.5 & 3.6 [3, 5] & 3.6 [3, 5]  & 3.6 [3, 4]  & 3.4 [3, 4]   \\
		1.0 & 1.8 [1, 2]  & 2 [2, 2]   & 2.8 [2, 3]  &  2.6 [2, 3] \\ \midrule
		\multicolumn{5}{l}{\textbf{Task: PII with target \fscore= 0.95, $E = 5$}} \\ \midrule
		0.05 & 39.6 [32, 44]  & 48.4 [35, 58]  & 34.6 [31, 40] & 39.2 [31, 44]  \\
		0.1 & 21.6 [16, 37]  & 20.2 [14, 27]  & 16.2 [14, 17]  & 20.2 [18, 22] \\
		0.2 & 9.4 [7, 14]  & 10.4 [8, 16]   & 7.8 [7, 8] &  9.2 [7, 11]\\
		0.5 & 3.2 [2, 5]  & 3.6 [3, 5] & 3 [3, 3]  & 3.2 [3, 4]]  \\
		1.0 & 1 [1, 1] & 1.2 [1, 2]  & 1.2 [1, 2]  & 1 [1, 1]  \\ \bottomrule
	\end{tabular}
	\caption{
	{\bf Results 2b.} Impact of \federated parameters. Consider the \nomoads dataset split into 20 synthetic users. All models are trained until they reach a target \fscore (selected to match \centralized for the same  task).  We vary the \federated parameters $C$, $B$, $E$ and we report the communication cost (number of rounds, $R$) until the target \fscore is reached: average and [min, max] are reported over 5 runs.   }
	\label{federated_table}
\end{table}



\textbf{Results 2a: \federated vs. \centralized vs. \standalone.} Table \ref{nomoads_pii_federated_vs_centralized} shows the classification performance (\fscore), where  the \federated model performs as well as the \centralized model and significantly outperforms the \standalone models. 
In particular, \federated training performs similarly to the same model  trained in  \centralized way on the union of all users' data. Moreover, the \fscore of the \federated model on each individual user's test data is slightly higher than their \standalone models, especially for the \uneven split. For \uneven split, the average number of rounds required to reach such \fscore for the Ads prediction is 8.8, 
while for PII prediction 1 round was sufficient to reach \fscore= 0.96. For \even split, the average number of rounds required to reach such \fscore is 2.6 for the Ads prediction  
 and 2.2 rounds for the PII prediction. 

\textbf{Setup 2b.} This  is similar to Setup 2a, but the data is split into 20, instead of 5, synthetic users. 
For B = $\infty$, we use all available data  without any minibatches, similarly to \cite{original_federated}. 
We require all models to reach a target \fscore on test set (0.85 and 0.95 for Ads and PII predictions respectively); we select these \fscore targets to match the performance of \centralized models shown in Table \ref{models} for each task.  

\textbf{Results 2b: Impact of \federated parameters.} Table \ref{federated_table} shows how the average number of rounds ($R$), until the models reach the target \fscore, depends on the  fraction of participating clients ($C$), a different batch size ($B$) and local epochs ($E$).  A general trend is that increasing the $C$ parameter, the average number of rounds decreases significantly and the gap between min and max decreases. Moreover, increasing the batch size decreases the number of rounds, as small batch size helps the model to converge faster. These observations apply to both \uneven and \even splits and to both prediction tasks. On the other hand, increasing the local epochs and pushing computation to users increases the number of rounds, except for the case when $C=1.0$. The reason for this is that our model is simple and more local epochs lead to overfitting. 
The main parameter that significantly affects the number of communication rounds  is $C$. This is expected and the effects of $C$ can be understood in the following way: using fewer clients in a round requires less communication per round, but similar amount of computation is required to train our models, hence the increased number of rounds.  
However, in the context of mobile packet classification, the number of rounds is much lower than observed for more complex models in related work \cite{McMahan2016} and remains around 10 for $C=0.2$.

\subsection{Scenario 3: AntShield for PII Prediction}

\textbf{Setup 3.} We use the following setup from Table \ref{all_options}. \textit{Dataset:} \antshield.  \textit{Users:} \Even vs. \Uneven split with 5 synthetic users. \textit{Classifier Granularity:} General.  \textit{Models:} \federated \svm vs. \centralized \svm, \textit{Tasks:} PII exposure. We set $B=10, E=5, C=1.0$, 
similarly to Setup 2.

\begin{table}[!t]
	\centering
	\small
	\begin{tabular}{llll}
		\toprule
		\multirow{2}{*}{\textbf{Trained on}} & \multirow{2}{*}{\textbf{Tested on}} & \textbf{\Uneven split}  & \textbf{\Even split}\\
		& &  \textbf{\fscore} & \textbf{\fscore}\\ \midrule
		\federated & user 0 test & 0.91 & 0.93  \\
		\federated & user 1 test & 0.94 & 0.95  \\
		\federated & user 2 test & 0.95 & 0.94  \\
		\federated & user 3 test & 0.95 & 0.91  \\
		\federated & user 4 test & 0.93 & 0.93  \\
		\federated & all test data & 0.93 & 0.93 \\
		\midrule
		\standalone user 0 & user 0 test & 0.92 & 0.89 \\
		\standalone user 1  & user 1 test & 0.91 &  0.91\\
		\standalone user 2  & user 2 test & 0.93 &  0.92 \\
		\standalone user 3  & user 3 test & 0.87 & 0.87 \\
		\standalone user 4  & user 4 test & 0.85 &  0.89\\
		\centralized & all test data & 0.94 & 0.94 \\ \bottomrule
	\end{tabular}
	\caption{ {\bf Results 3.} \antshield dataset for predicting PII exposures, for 5 synthetic users created with \uneven and \even split of data. The \fscore is averaged from 5 runs for $C=1.0, B=10, E=5$.} 
	\label{antshield_5_users_equal_vs_skewed}
\end{table}

\textbf{Results 3.} Table \ref{antshield_5_users_equal_vs_skewed} shows the results. 
For \even split of data, the \federated model has an \fscore of 0.94 when it is tested on the union of user test sets, while the corresponding \centralized model has an \fscore= 0.96, achieved within 5.8 rounds on average. 
 For the \uneven split of data among users, the \federated model achieves the same \fscore= 0.94, but slightly slower (in 6.6 rounds). 
We observe that some users achieve lower \fscore on their corresponding \standalone models, which is expected as these users have much less data and especially positive examples, because of the skewness of data in the \uneven split. In summary, we show that even with a different dataset, our \federated approach still performs well when compared to its corresponding \centralized model for both types of splits, with a small difference in communication rounds to achieve the same \fscore. 

\subsection{Scenario 4: In-house Datasets for PII Prediction}\label{sec:inhouse}

\begin{figure}[!t]
	\centering
	\includegraphics[width=1.0\linewidth]{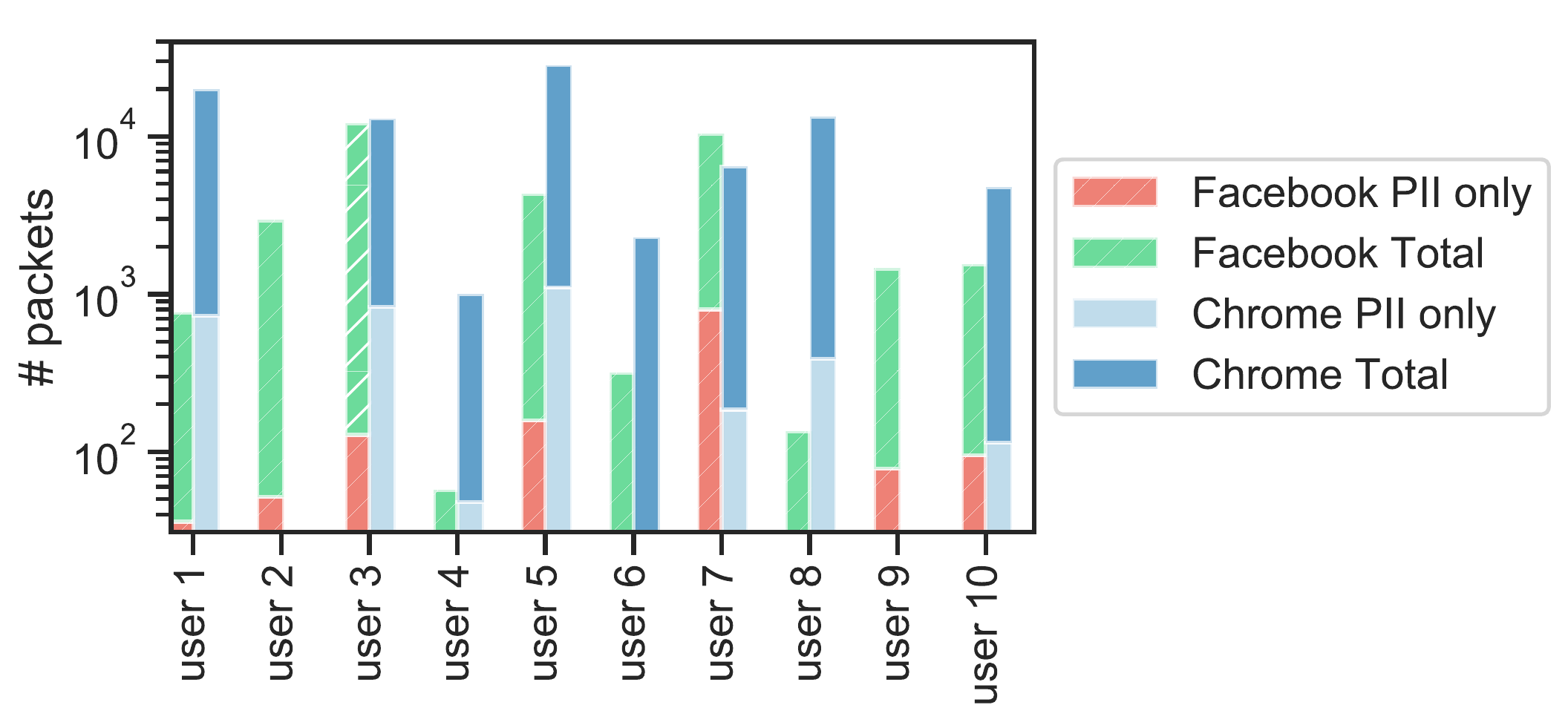}
	\caption{Distribution of packets  for Facebook and Chrome.}
	\label{fig:chrome_distribution}
\end{figure}

\textbf{Setup 4.} We use the following setup from Table \ref{all_options}: \textit{Dataset:} \Inhouse Chrome, Facebook. \textit{Users:} 10 real users. \textit{Classifier Granularity:} Per App. \textit{Models:} \federated \svm vs. \centralized \svm. \textit{Tasks:} PII exposure.  The goal is  to evaluate  our \federated framework (1) on real user activity (instead of systematic tests of apps) and (2) over a longer time period (7 months instead of five/ten minutes).
Figure \ref{fig:chrome_distribution} shows the distribution of Chrome and Facebook packets (including labels) present across the 10 real users in our \inhouse dataset.


\textbf{Results 4a.}  Table \ref{minas_summary} shows the classification performance of the \centralized and \federated models for Chrome and Facebook, with certain parameters. 
 Chrome's \federated model achieves \fscore= 0.84 compared to its \centralized \fscore= 0.92. Facebook's \federated model maintains similar \fscore (0.94) compared to its \centralized version (0.95).



\begin{table}[!t]
	\setlength\tabcolsep{4pt} 
	\centering
	\small
\begin{tabular}{lllll}
	\toprule
	 & \multicolumn{2}{l}{\textbf{Chrome}} & \multicolumn{2}{l}{\textbf{Facebook}} \\ \midrule
	 \textbf{Model} & \fscore & R {[}min, max{]} & \fscore & R {[}min, max{]} \\ \midrule
	\federated & 0.84 & 94 [9, 212]  & 0.94 & 33.2 [5, 113]  \\
	\centralized & 0.92 & N/A & 0.95 &  N/A\\ \bottomrule
\end{tabular}
	\caption{{\bf Results 4a.} PII prediction for 
	Chrome, 
    Facebook.  
	  We report the average \fscore and 
	 the number of rounds, $R$ (avg [min, max]) required to achieve that \fscore. We set $C=0.5, B=10, E=5$.}
	\label{minas_summary}
\end{table}

\begin{table}[!t]
	\centering
	\small 
	\begin{tabular}{llll}
		\toprule
		$E$ &$B$ & \multicolumn{2}{l}{R: avg [min, max]} \\  \midrule 
		\multicolumn{2}{l}{}& \textbf{Facebook} & \textbf{Chrome} \\ \midrule
		1 & 10 &  16 [6, 31] & 7  [4, 10]  \\
		1 & 20 & 20.2 [12, 41]  & 6.4 [5, 11]  \\
		1 & 40 & 15.6 [8, 27] & 5 [4, 7]] \\
		1 & $\infty$ & 9 [6, 14]  & 6.2 [4, 11]  \\
		5 & 10 & 33.2 [5, 113]   & 82 [14, 200] \\
		5 & 20 & 37.6 [20, 46] & 27 [3, 61]  \\
		5 & 40 & 39.6 [8, 97]  & 25.6 [8, 55] \\
		5 & $\infty$ & 26 [3, 47]  & 23.2 [6, 56]  \\
		10 & 10 & 53.8 [4, 190] &  756.2 [531, 800] \\
		10 & 20 & 71.6 [12, 200]  &  - \\ 
		10 & $\infty$ & 72.8 [21, 146] &  283.2 [126, 800] \\ \bottomrule
	\end{tabular}
	\caption{ {\bf Results 4b.} We report the average [minimum, maximum] number of communication rounds ($R$) required to achieve a target \fscore of 0.94 and 0.84  for Facebook and Chrome, respectively. We vary the parameters batch size ($B$) and number of local epochs ($E$) to evaluate their impact on the average number of rounds, with fixed $C=0.5$. 
	If the target \fscore is not reached within 800 rounds for none of the 5 runs, we assume that it does not converge.}
	\label{B_E_vs_rounds}
\end{table}

\textbf{Results 4b.} In Table \ref{B_E_vs_rounds}, we evaluate the impact of batch size ($B$) and local epochs ($E$) on the communication  rounds required to achieve a target \fscore for Chrome and Facebook.  We run each experiment five times in order to obtain the average number of rounds required and the min, max values. We observe that increasing the batch size increases slightly the average number of rounds to achieve the target \fscore, while increasing the $E$ parameter increases the number of rounds significantly. The reason is that we use a simple model 
and most likely the model overfits with large epoch values. In the original Federated Learning paper \cite{original_federated}, the authors  showed the opposite effect: increasing the local epochs decreases the number of rounds; however, they train DNNs that require more complexity and do not overfit for those $E$ values. Moreover, we observe that the batch size ($B$) does not significantly affect the number of rounds required.  
However, the  number of local epochs ($E$)  plays an important role in the model's convergence, which we further explore next. 

\begin{figure}[!t]
	\centering
	\begin{subfigure}{0.5\textwidth}
		\centering
		\includegraphics[width=0.95\linewidth]{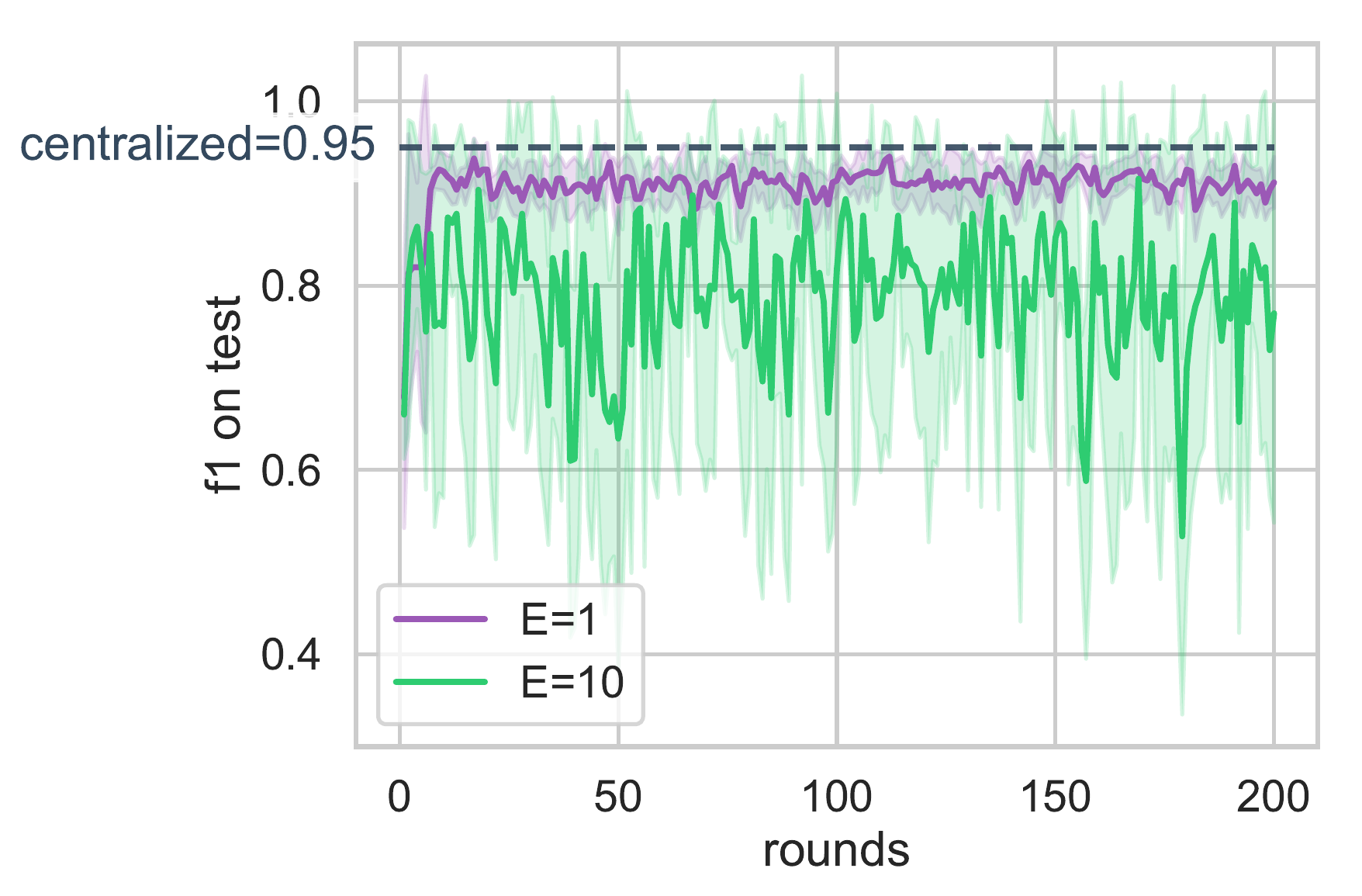}
		\caption{Facebook classifier; rounds vs. local epochs $E$.}
		\label{fig:facebook_vs_epoch}
	\end{subfigure}
	\\
	\begin{subfigure}{0.5\textwidth}
	\centering
	\includegraphics[width=0.95\linewidth]{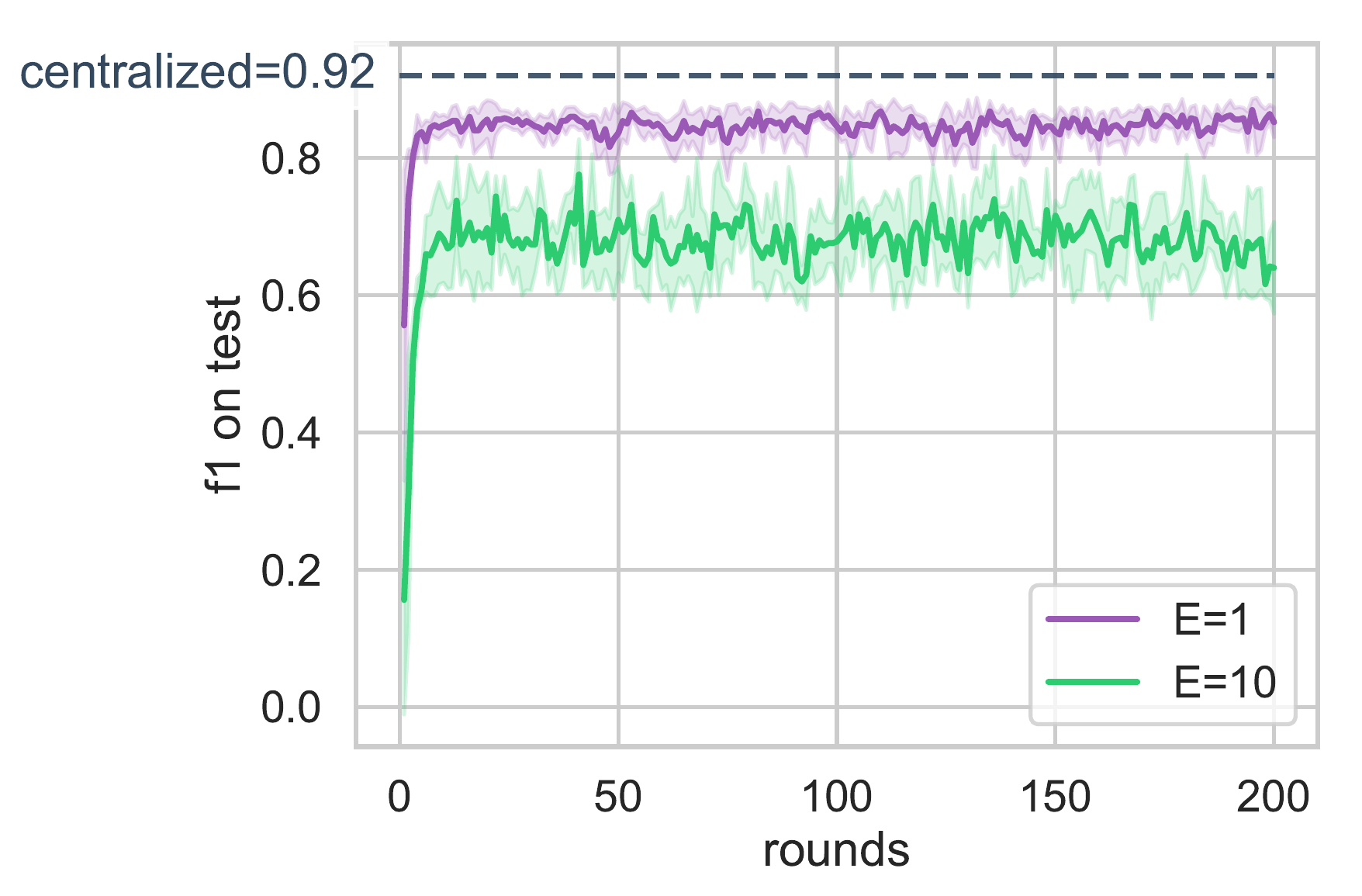}
	\caption{Chrome classifier; rounds vs. local epochs $E$.}
	\label{fig:chrome_vs_E}
	\end{subfigure}
	\caption{{\bf Results 4c.} Convergence of \fscore over $R$ communication rounds for the Facebook and Chrome \federated classifiers. We fix $C=0.5, B=10$ and vary $E$. Models are trained 5 times, and shaded regions represent the standard deviation from the average \fscore. Their \centralized models reach \fscore 0.95 and 0.92 (dashed line). } 
	\label{fig:chrome_facebook}
\end{figure}

\textbf{Results 4c: Convergence of \federated models.} Figure \ref{fig:chrome_facebook} shows the performance of \federated \svm for Facebook and Chrome when we vary the local epochs, $E$. 
We set the fraction of client $C=0.5$ and batch size $B=10$. We train each model with an $E$ value five times and report the average and standard deviation (in shadowed color). The main difference between the two apps is that the \fscore of the  \federated model is closer to the \centralized one for Facebook. However, the standard deviation is much larger than Chrome's. In addition, $E=1$ for Chrome can reach a better \fscore= 0.89 than in the previous experiments, because of the lower $E$ value. 
We observe that the \federated model is more sensitive to the $E$ parameter, which leads to overfitting for \svm. 

\begin{figure}[!t]
	\centering
	\includegraphics[width=0.92\linewidth]{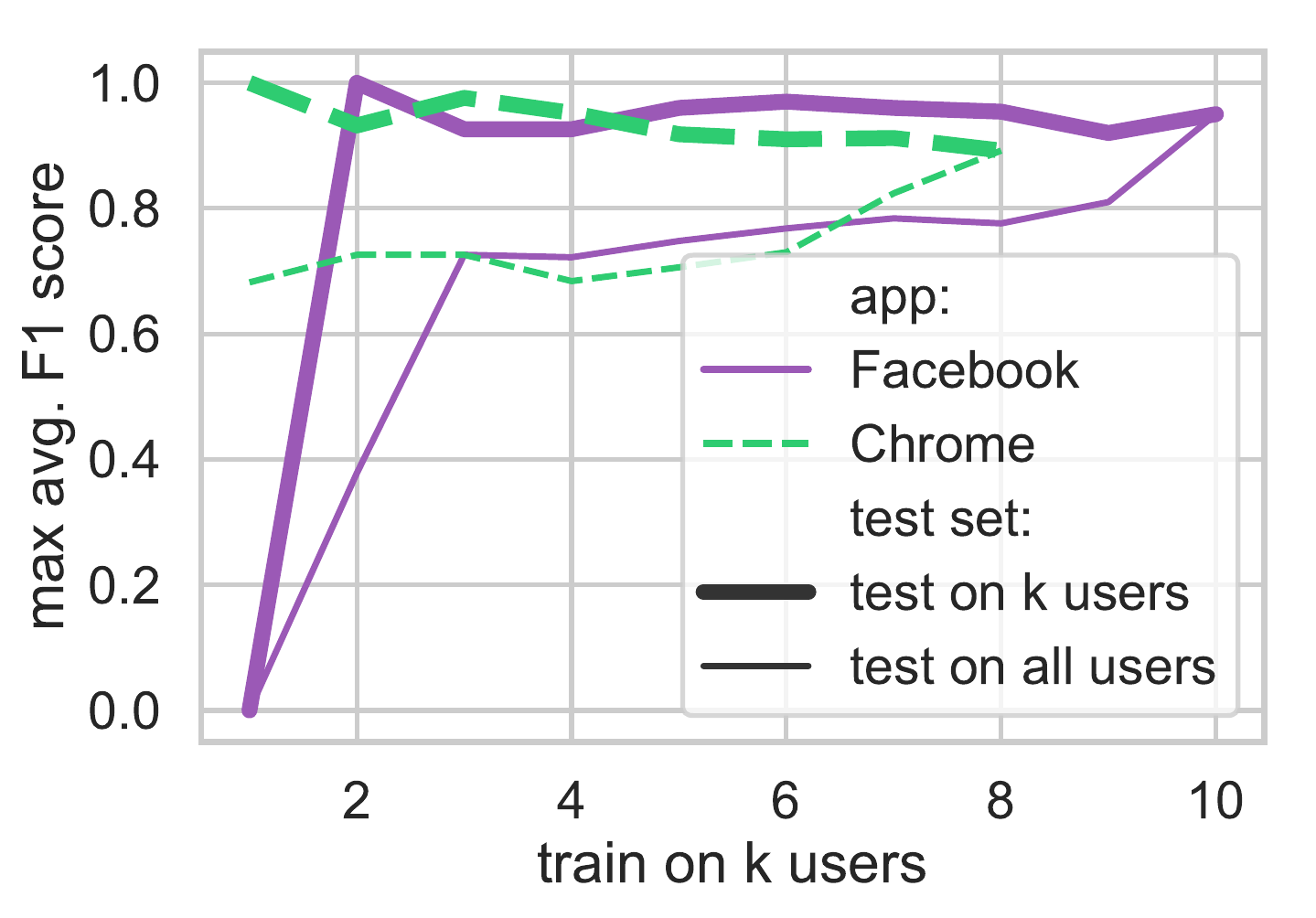}
	\caption{{\bf Results 4d.} Benefit of crowdsourcing for Chrome and Facebook, where k users participate in training. We sort users in increasing amount of training data they contribute. 
	Federated parameters: $C=1.0, E=5, B=10$. The average \fscore 
	is shown for  
	all users' test data (thin line) and for test data of k users (bold line). } 
	\label{fig:chrome_vs_fb_benefit_crowd_max_f1}
\end{figure}

\textbf{Results 4d: Benefit of Crowdsourcing.} We ask the question: how many users need to collaborate in training a global model in order to get most of the predictive power? 
Figure \ref{fig:chrome_vs_fb_benefit_crowd_max_f1} shows that a few users participating in the training phase during \federated learning can be beneficial for all users. 
We show the maximum average \fscore obtained from five runs, as a function of the number of users  ($k$)  participating in  training. The \fscore is evaluated both on all user's test data and on the test data of $k$ users who participated in the training. We sort the users by increasing amount of  training data. For example, when $k=1$, one user with the fewest data points participates in training. 
When $k=2$, in addition to the previous user, another user with more data is used during training. 
More users in the training phase is beneficial to increase the \fscore for both apps. However, there are some users who do not help with their data and slightly worsen the \fscore, as their data might confuse the classifier. For Facebook, we need to add at least 3 users in order to obtain a decent \fscore, while Chrome reaches the same \fscore with only 2 users. 
The \fscore on the test data of k users is much higher than on the union of all users' test data, as the models only fit to the data available for the participating users. 
The lack of generalization is one of the reasons that \webview apps are a challenging special case in the mobile packet classification problem. However, both Chrome and Facebook train \federated models that generalize well with \fscore of over 0.80, if enough users with useful (diverse) data participate in training. 

\subsection{Scenario 5: Interpreting SVM vs. DT}\label{sec:interpret}

\textbf{Setup 5.} We use the following setup from Table \ref{all_options}: \textit{Dataset:} \nomoads.  \textit{Users:} None. \textit{Classifier Granularity:} General. \textit{Models:} \centralized \svm vs. DT. 
\textit{Tasks:} PII exposure.
Prior work chose DT over other models partially because of their interpretability. 
 In our context, these models learn similar separation of our datasets, which we demonstrate by (1) observing the most important coefficients in \svm, (2) by knowledge transfer from \svm to DT. 
The goal here is to compare \svm to DT in terms of their interpretability.


\begin{figure}[!t]
	\centering
	\includegraphics[width=0.89\linewidth]{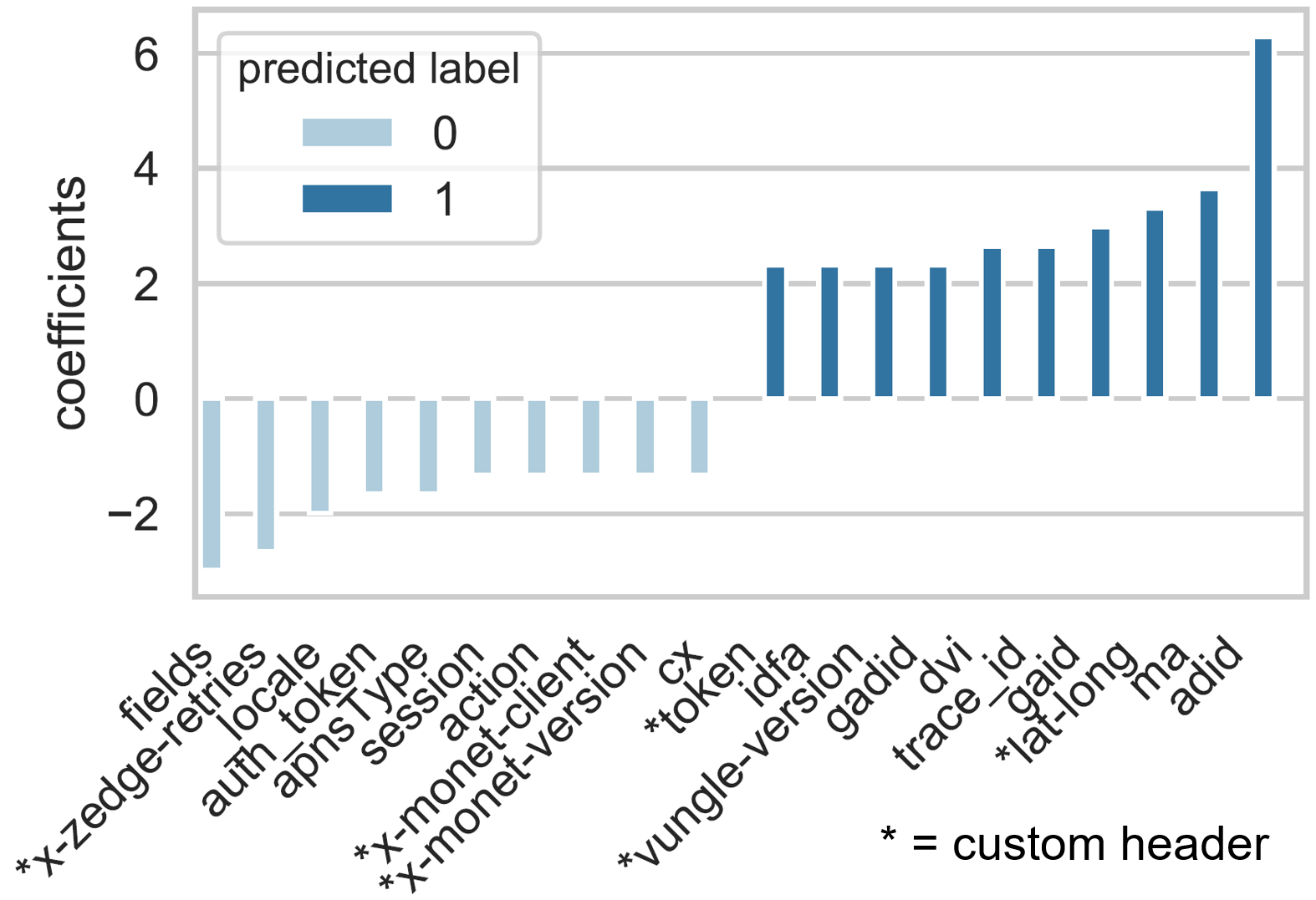}
	\caption{Top 10 negative and positive coefficients and the corresponding features obtained from \centralized \svm.} 
	\label{fig:nomoads_pii_svm}		
\end{figure}

\begin{figure}[!t]
	\begin{subfigure}{0.5\textwidth}
	\centering
	\includegraphics[width=0.95\textwidth]{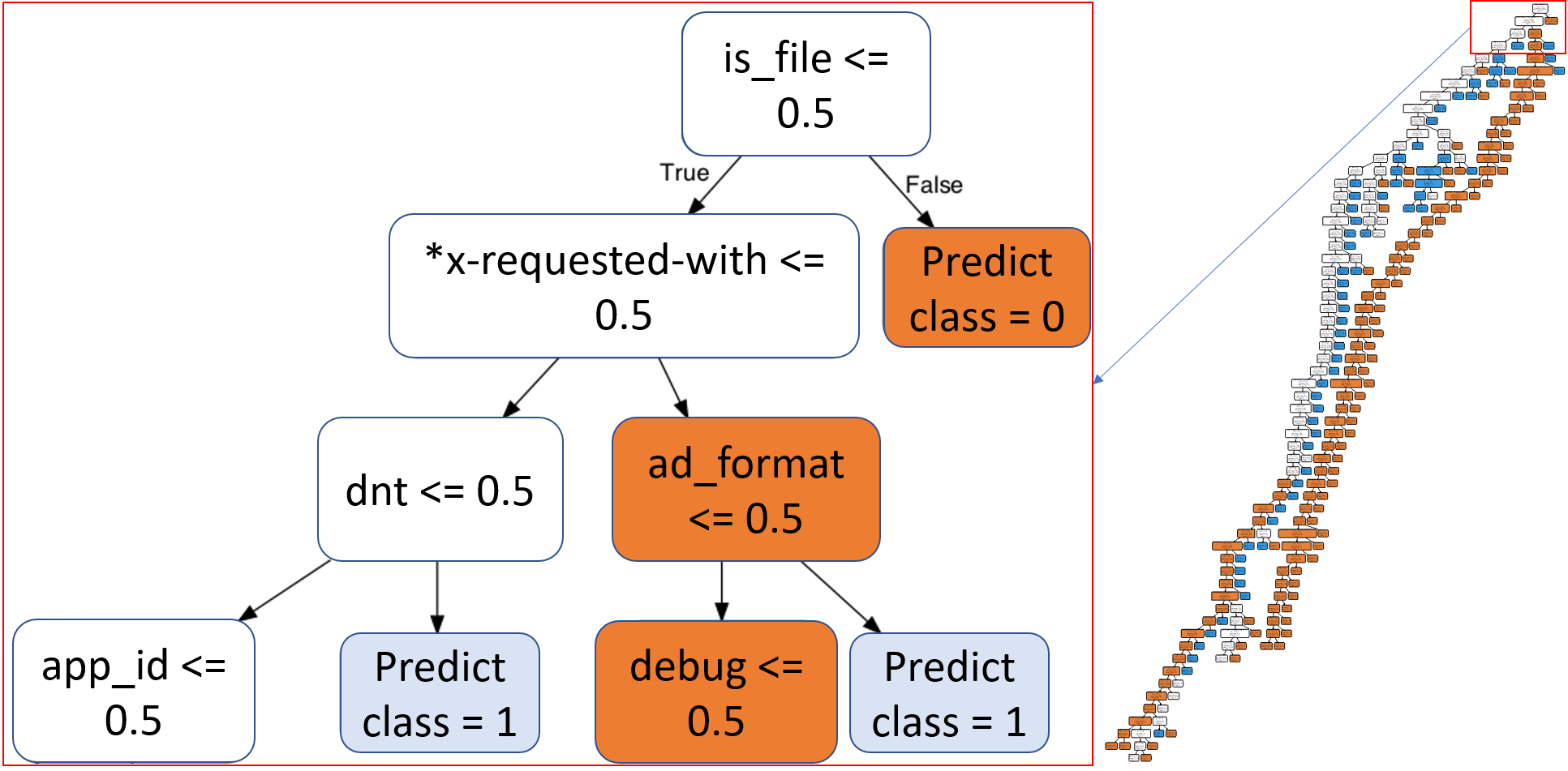}
	\caption{Decision Tree (DT) trained on its own.}
	\label{fig:nomoads_pii_original_tree}
\end{subfigure}
\begin{subfigure}{0.5\textwidth}
	\centering
	\includegraphics[width=0.95\textwidth]{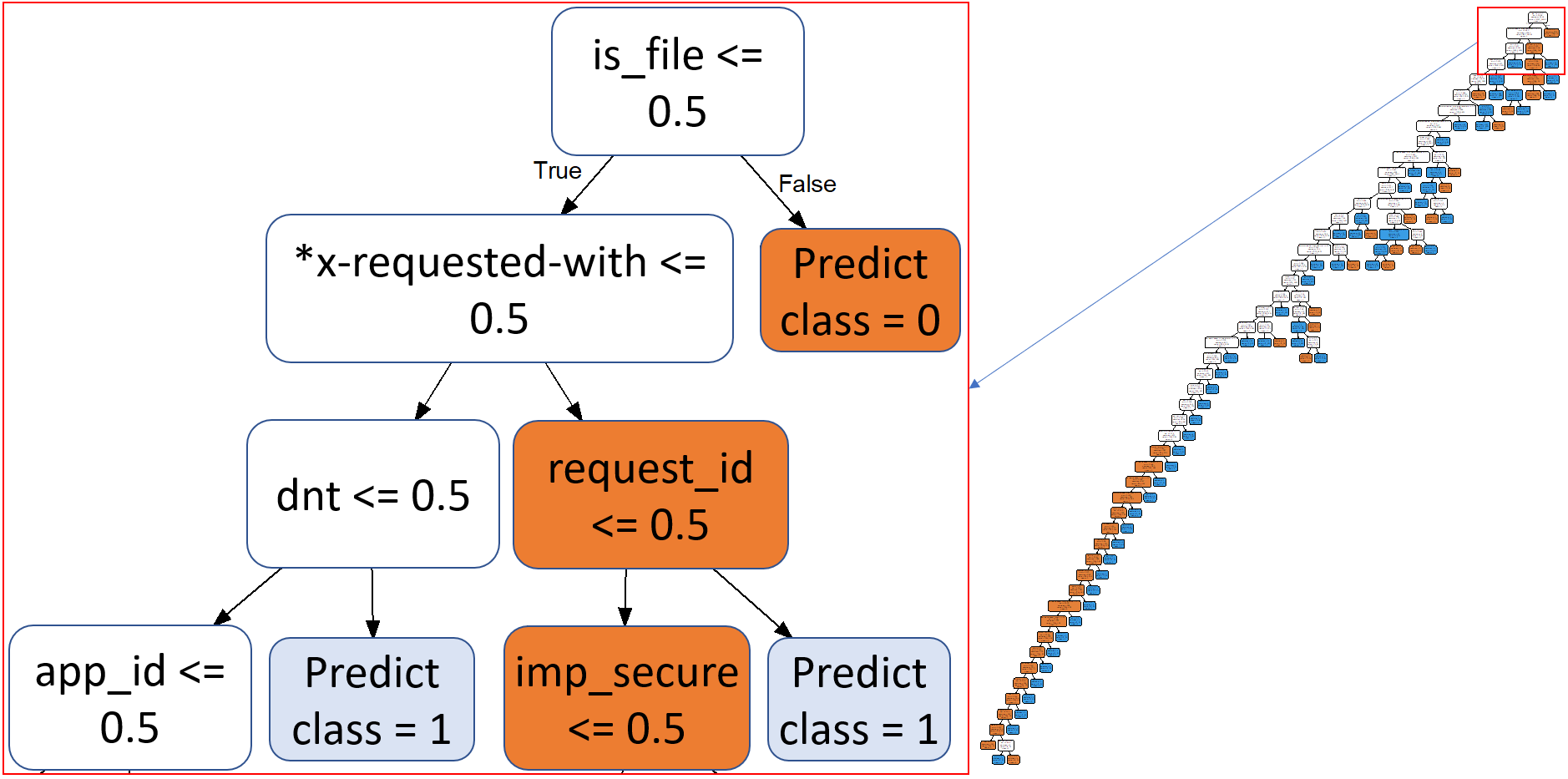}
	\caption{First train \svm, then transfer knowledge to DT.}
	\label{fig:nomoads_pii_from_svm_tree}
\end{subfigure}
\caption{Interpretability of DT vs. \svm in Setup 5. Due to lack of space, we show zoomed-in versions of these DTs in Appendix \ref{sec:zoomin}.}
\label{fig:knowledge_transfer}
\end{figure}

\textbf{Results 5.}
Figure \ref{fig:nomoads_pii_svm}  presents the ten most important negative and positive coefficients and their corresponding features for our \centralized \svm model. 
In order to distinguish important features, we use the model's coefficients, where the positive ones correspond to the features whose presence leads to positive labels  and the negative coefficients correspond to features responsible for predicting label 0 (\eg No PII detected).  This is not a one-to-one mapping of important features between \svm and DT due to their internal representation of features. However, we observe certain keys that are responsible for PII exposures such as ``gaid'', that also appear in the corresponding Decision Tree. 

Figure 
\ref{fig:knowledge_transfer}(\subref{fig:nomoads_pii_from_svm_tree}) shows the DT after knowledge transfer from \svm.
To perform knowledge transfer from \svm to DT, we first split the data into 40\% for training \svm, another 40\% for training a DT, which is labeled with predictions from the aforementioned \svm. The remaining 20\% of the data is used for testing. 
 This is one way to leverage the interpretability of Decision Trees via knowledge transfer from \svm. In Fig. \ref{fig:knowledge_transfer}(\subref{fig:nomoads_pii_original_tree}),
 we show a 
 Decision Tree which was trained with \nomoads for PII prediction, while in Fig. \ref{fig:knowledge_transfer}(\subref{fig:nomoads_pii_from_svm_tree}),
 we show the DT after knowledge transfer from \svm. We observe that both DTs, at least at the top levels, have similar important features and thus, capture similar patterns. 
The original DT and \svm reached \fscore= 0.95 and the after knowledge transfer DT reached \fscore= 0.94 on the same test data. 
This is only a minor \fscore loss during knowledge transfer.

The most notable difference between the trees in Fig. \ref{fig:knowledge_transfer} is the lack of a large branch that only predicts label 0, which is the result of how the original tree unsuccessfully attempts to separate data. However, the DT after the knowledge transfer is oblivious to this error, since the \svm most likely suffers from the same issue as the original DT. Such errors propagating from the \svm make the DT after the knowledge transfer  smaller (269 vs. 141 nodes) than the original DT.

\section{Conclusion and Future Directions}\label{sec:conclusion}

\textbf{Summary.} This paper proposes, for the first time, a Federated Learning framework for Mobile Packet Classification, and evaluates  its effectiveness and efficiency, using three real-world datasets and  two different tasks (namely PII exposure and Ad request). First, we proposed a reduced feature space (\httpkeys), which limits the sensitive information shared by users. 
 Then, we showed that \svm with SGD performs similarly to Decision Trees used by state-of-the-art \cite{recon15, antshield-arxiv, spawc}, in terms of \fscore as well as interpretability. We also 
 showed that \federated achieves a significantly higher \fscore than \standalone and is comparable to \centralized models, and it does so 
 within a few communication rounds and with minimal computation per user, which is important in the mobile environment. 
 
\textbf{Future work.} There are several directions for future work. First, we plan to study the exploding feature space of \webview apps   
due to temporal and user dynamics. Second, we will seek to evaluate our framework in larger datasets, which would also allows us to train and evaluate DNNs. Third, we will explore ways to handle encrypted traffic (\eg by focusing on network features, SNI, domain embeddings, etc). Fourth, we will consider well-known privacy attacks to Federated Learning (model parameter updates can still contain sensitive information) and mechanisms (Secure Aggregation \cite{Bonawitz} or Differential Privacy \cite{dwork2011differential})  that can be added on top of Federated Learning, which were considered out of scope for this paper. In future work, we are interested in optimizing the design of such  mechanisms specifically for the mobile classification problem. Fifth,  we will  leverage knowledge transfer from SVMs to deploy DT classifiers on mobile, which have been previously implemented efficiently on the device \cite{antshield-arxiv}, thanks to the simple DT rules. 
Related to that, we plan to implement our method directly on mobile phones and possibly leverage the recently announced open source framework for Federated Learning, TensorFlow Federated \cite{tf-federated}, which can be done on top of open source VPN tools \cite{antshield-arxiv, nomoads}. 
A mobile implementation would provide us with evaluation on real-world resource consumption. 
Finally, our general framework can be applied towards other packet classification tasks, beyond PII and Ad prediction, such as \eg fingerprinting, provided there are labeled packet traces to train on.

\section{Acknowledgments}\label{sec:acknowledgements}

This work has been partially funded by NSF Awards 1649372 and 1526736. E. Bakopoulou and B. Tillman have both been partially supported by H. Samueli and Networked Systems Fellowships. E. Bakopoulou has also received a Broadcom Foundation Fellowship.

\bibliographystyle{unsrt}
\bibliography{allrefs_sorted}
\newpage
\appendix

\section{Additional Figures} \label{sec:zoomin}

\noindent
\begin{minipage}{\textwidth}
    \centering
    
    \includegraphics[height=0.9\textheight]{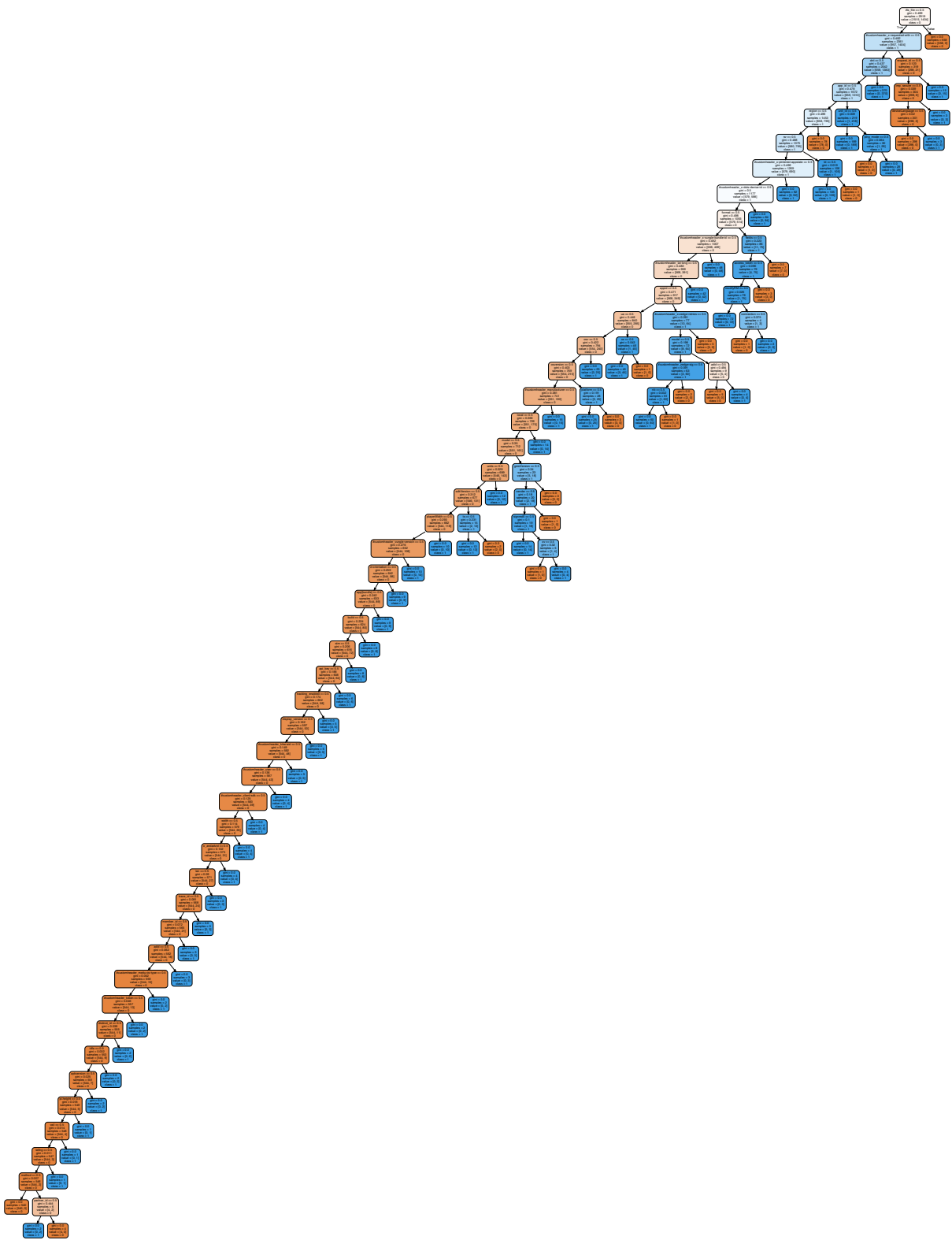}
    \captionof{figure}{Zoomed-in version of Fig. \ref{fig:knowledge_transfer}(\subref{fig:nomoads_pii_from_svm_tree}), showing a Decision Tree (DT) after knowledge transfer from an \svm.}
    \label{fig:fig:nomoads_pii_from_svm_tree_full}
\end{minipage}


\begin{figure*}[!t]
	\centering
	\includegraphics[height=0.99\textheight]{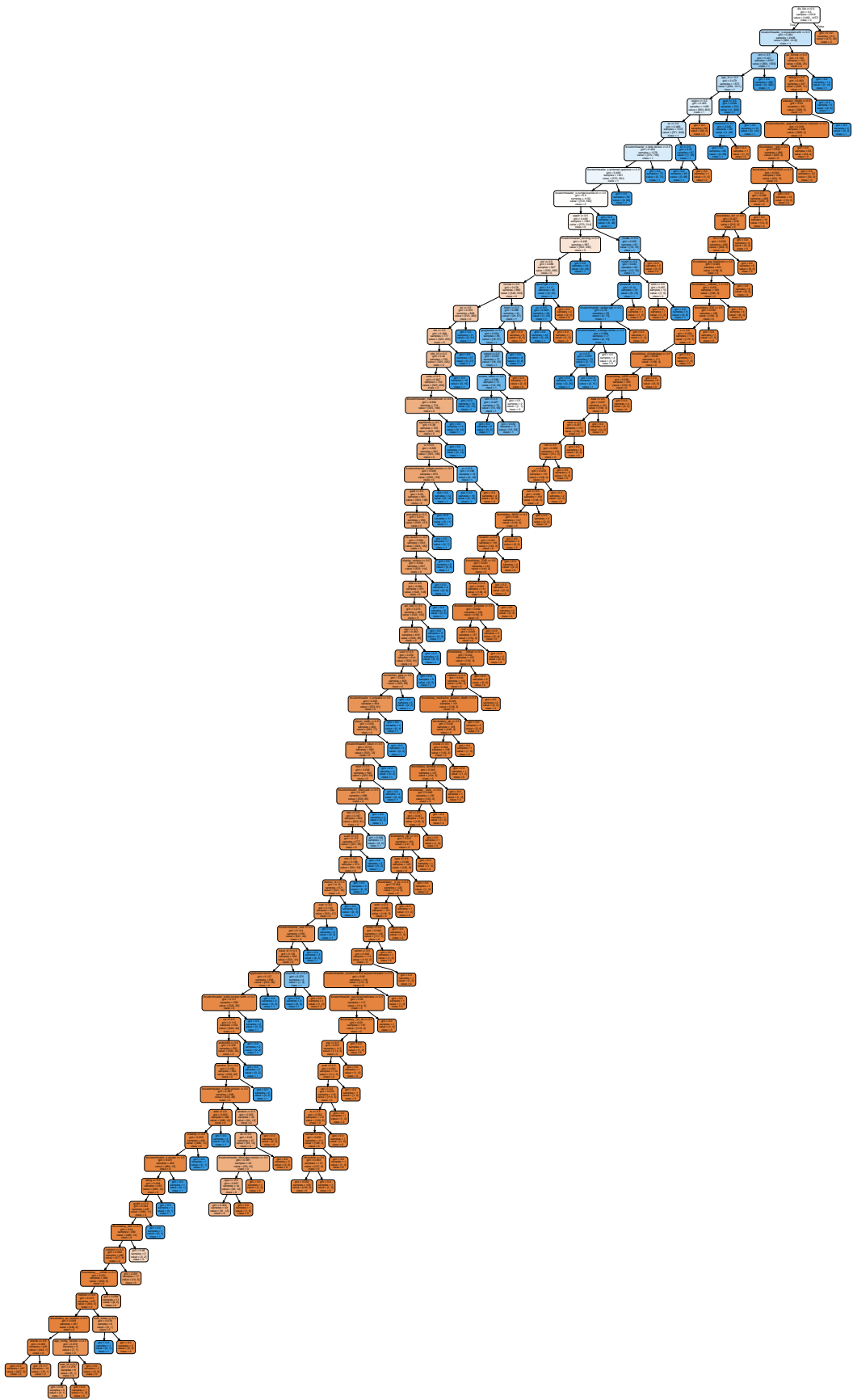}
	\caption{Zoomed-in version of Fig. \ref{fig:knowledge_transfer}(\subref{fig:nomoads_pii_original_tree}), showing a Decision Tree (DT) trained on its own.}
	\label{fig:fig:nomoads_pii_original_tree_full}
\end{figure*}
\newpage

\end{document}